\documentclass[unnumsec,webpdf,traditional,large,namedate]{oup-authoring-template}

\usepackage{etoolbox}
\makeatletter
\patchcmd{\@maketitle}{\color{black!20}\rule{45pt}{55pt}}{\color{white}\rule{45pt}{55pt}}{}{}
\patchcmd{\@maketitle}{\color{black!20}\rule{45pt}{55pt}}{\color{white}\rule{45pt}{55pt}}{}{}
\patchcmd{\@maketitle}{\color{black!20}\rule{45pt}{55pt}}{\color{white}\rule{45pt}{55pt}}{}{}
\patchcmd{\@maketitle}{\color{black!20}\rule{45pt}{55pt}}{\color{white}\rule{45pt}{55pt}}{}{}
\patchcmd{\@maketitle}{\color{black!20}\rule{45pt}{55pt}}{\color{white}\rule{45pt}{55pt}}{}{}
\patchcmd{\@maketitle}{\color{black!20}\rule{45pt}{55pt}}{\color{white}\rule{45pt}{55pt}}{}{}
\patchcmd{\@maketitle}{\color{black!20}\rule{45pt}{55pt}}{\color{white}\rule{45pt}{55pt}}{}{}
\patchcmd{\@maketitle}{\color{black!20}\rule{45pt}{55pt}}{\color{white}\rule{45pt}{55pt}}{}{}
\patchcmd{\@maketitle}{\color{black!20}\rule{45pt}{55pt}}{\color{white}\rule{45pt}{55pt}}{}{}
\makeatother

\onecolumn 

\graphicspath{{Fig/}}

\usepackage{graphicx}%
\usepackage{multirow}%
\usepackage{amsmath,amssymb,amsfonts}%
\usepackage{amsthm}%
\usepackage{mathrsfs}%
\usepackage{xcolor}%

\usepackage{bm}
\usepackage{enumerate}
\usepackage{caption}
\usepackage{color}
\usepackage{comment}
\usepackage{mathtools}
\usepackage{dsfont}
\usepackage{float}

\usepackage{setspace}
\usepackage{longtable}
\usepackage{titlesec}

\titlespacing*{\section}
{0pt}{.4ex}{.4ex}
\titlespacing*{\subsection}
{0pt}{.4ex}{.4ex}

\newcommand{\mat}[1]{\bm{#1}} 
\let\vec\bm

\theoremstyle{thmstyleone}%
\newtheorem{theorem}{Theorem}
\newtheorem{proposition}[theorem]{Proposition}%
\newtheorem{corollary}{Corollary}
\theoremstyle{thmstyletwo}%
\theoremstyle{thmstylethree}%

\begin{document}

\journaltitle{}
\DOI{}
\copyrightyear{}

\firstpage{1}


\title[Merging versus Ensembling]{Merging versus Ensembling in Multi-Study Prediction: Theoretical Insight from Random Effects}

\author[1,$\ast$]{Zoe Guan}
\author[2, 3]{Giovanni Parmigiani}
\author[4]{Prasad Patil}

\authormark{Author Name et al.}

\address[1]{\orgdiv{Department of Biostatistics}, \orgname{Massachusetts General Hospital}}
\address[2]{\orgdiv{Department of Data Science}, \orgname{Dana Farber Cancer Institute}, \orgaddress{Boston, USA}}
\address[3]{\orgdiv{Department of Biostatistics}, \orgname{Harvard T.H. Chan School of Public Health}, \orgaddress{Boston, USA}}
\address[4]{\orgdiv{Department of Biostatistics}, \orgname{Boston University School of Public Health}, \orgaddress{Boston, USA}}

\corresp[$\ast$]{Corresponding author. \href{email:email-id.com}{zoe.guan@mgh.harvard.edu}}



\abstract{A critical decision point when training predictors using multiple studies is whether studies should be combined or treated separately. We compare two multi-study prediction approaches in the presence of potential heterogeneity in predictor-outcome relationships across datasets: 1) merging all of the datasets and training a single learner, and 2) multi-study ensembling, which involves training a separate learner on each dataset and combining the predictions resulting from each learner. For ridge regression, we show analytically and confirm via simulation that merging yields lower prediction error than ensembling when the predictor-outcome relationships are relatively homogeneous across studies. However, as cross-study heterogeneity increases, there exists a transition point beyond which ensembling outperforms merging. We provide analytic expressions for the transition point in various scenarios, study asymptotic properties, and illustrate how transition point theory can be used for deciding when studies should be combined with an application from metagenomics.}


\maketitle

\setstretch{2.02}

\vspace{-1cm}
\section{Introduction}
Prediction models trained on a single study often perform considerably worse in external validation than in cross-validation \citep{castaldi2011empirical, Bernau2014}. Their generalizability is compromised by overfitting, but also by various sources of study heterogeneity, including differences in study design, data collection and measurement methods, unmeasured confounders, and study-specific sample characteristics \citep{doi:10.1093/biostatistics/kxy044}. Using multiple training studies can potentially address these challenges and lead to more replicable prediction models. In many settings, such as precision medicine, multi-study prediction is motivated by systematic data sharing and data curation initiatives. For example, the establishment of gene expression databases such as Gene Expression Omnibus \citep{edgar2002gene} and ArrayExpress \citep{parkinson2010arrayexpress} and neuro-imaging databases such as OpenNeuro \citep{gorgolewski2017openneuro} has facilitated access to sets of studies measuring the same outcome and predictors. Such studies can be jointly analyzed \citep{Riester:2012gt, de2021bayesian, huttenhower2006scalable, rhodes2002meta}. 

When multiple studies are available, it is important to systematically integrate information across datasets. There are many preprocessing methods for standardizing data across studies to make them more comparable, especially to adjust for technical differences \citep{lazar2012batch, benito2004adjustment, johnson2007adjusting, zhang2020combat}. Scenarios where variables overlap but differ across studies, leading to systematically missing data \citep{resche2013multiple}, can be more challenging, although in some applications they can handled by methods such as multiple imputation \citep{audigier2018multiple}. 

One common approach is to training prediction models on multiple datasets is to merge them and treat all the observations as if they are from the same study \citep{xu2008merging, jiang2004joint, gupta2020predictive}. The resulting increase in sample size can lead to better performance when the datasets are relatively homogeneous. Also, the merged dataset is often representative of a broader reference population than any of the individual datasets. \cite{xu2008merging} showed that a prognostic test for breast cancer metastases developed from merged data performed better than tests developed using individual studies. \cite{zhou2017can} proposed hypothesis tests for determining when it is beneficial to pool data across multiple sites for linear regression, compared to using data from a single site. 

Another approach is to combine results from separately trained models. Meta-analysis and ensembling both fall under this approach. Meta-analysis combines summary measures from multiple studies to increase statistical power \citep{Riester:2012gt, tseng2012comprehensive,rashid:2019}. A common strategy is to take a weighted average of the study-specific summary measures. In fixed effects meta-analysis, the weights are based on the assumption that there is a single true parameter underlying all of the studies, while in random effects meta-analysis, the weights are based on a model where the true parameter varies across studies according to a probability distribution. When learners are indexed by a finite number of common parameters, meta-analysis applied to these parameters can be used for multi-study prediction, leading to improved prediction accuracy \citep{Riester:2012gt}. Various studies have compared meta-analysis to merging. For effect size estimation, \cite{bravata2001simple} showed that merging heterogeneous datasets can lead to spurious results while meta-analysis protects against such problematic effects. \cite{taminau2014comparison} and \cite{kosch2018conducting} found that merging with batch effect removal had higher sensitivity than meta-analysis in gene expression analysis, while \cite{lagani2016comparative} found that the two approaches performed comparably in reconstruction of gene interaction networks. Ensemble learning methods \citep{dietterich2000ensemble}, which combine predictions from multiple models, can also be used to leverage information from multiple studies. Ensembling can lead to lower variance and higher accuracy, and is applicable to more general classes of learners than meta-analysis. \cite{patil2018training} proposed multi-study ensembles, which are weighted averages of prediction models trained on different studies, as an alternative to merging. They showed empirically that when the datasets are heterogeneous, multi-study ensembling can lead to improved generalizability and replicability compared to merging and meta-analysis.  

In this paper, we provide explicit conditions under which merging outperforms multi-study ensembling and vice versa. We study merged and ensemble learners based on ridge regression, allowing for basis expansions of predictors to capture a flexible and general class of linear models. Using a flexible mixed effects model to represent heterogeneity, we show that merging has lower prediction error than ensembling when heterogeneity is low, but as heterogeneity increases, there exists a transition point beyond which ensembling outperforms merging. We characterize this transition point analytically under a fixed weighting scheme for the ensemble. We also discuss optimal ensemble weights for ridge regression. We study the transition point via simulations and compare merging and ensembling in practice, using microbiome data.

\section{Problem Definition}

\subsection{General Notation}

We will use the following matrix notation: $\mat{I_N}$ is the $N \times N$ identity matrix, $\mat{0_{N \times M}}$ is an $N \times M$ matrix of 0's, $\mat{0_{N}}$ is a vector of 0's of length $N$, $tr(\mat{M})$ is the trace of matrix $\mat M$, $diag(\vec{u})$ is a diagonal matrix with $\vec{u}$ along its diagonal, and $(\mat M)_{ij}$ is the entry in row $i$ and column $j$ of matrix $\mat M$. Other notation used in the paper is summarized in Appendix D (Supplementary Materials).

\subsection{Data Generating Model}

Suppose we have $K$ studies that measure the same outcome and the same $P$ predictors, and the datasets have been harmonized so that measurements across studies are on the same scale. For individual $j$ in study $k$, let $Y_{jk}$ be the outcome and $\vec{X_{jk}} \in \mathds{R}^P$ the vector of predictors. Let $X_{ijk}$ be the value of the $i$th predictor for individual $j$ in study $k$. Let $N_k$ be the number of individuals in study $k$, $\vec{Y_k} = (Y_{1k}, \dots, Y_{N_k k}) \in \mathds{R}^{N_k}$, and $\mat{X_k}=\begin{bmatrix}\mat{X_{1k}} | \dots | \mat{X_{N_k k}}\end{bmatrix}^T  \in \mathds{R}^{N_k \times P}$. Assume the data are generated from the mixed effects model
\begin{flalign}\label{model1}
\vec{Y_{k}} = \mat f(\mat{X_{k}}) + \mat{Z_{k}}\vec{\gamma_k} + \vec{\epsilon_{k}}
\end{flalign}
where $\mat f(\mat{X_{k}}) = \begin{bmatrix} f_0\left(\mat{X_{1k}}\right) | \dots | f_0\left(\mat{X_{N_k k}} \right) \end{bmatrix}^T$ for some arbitrary function $f_0: \mathds{R}^{P} \to \mathds{R}$, $\mat{Z_{k}} \in \mathds{R}^{N_k \times Q}$ is the design matrix for the random effects, $\vec{\gamma_k} \in \mathds{R}^{Q}$ is the vector of random effects with $E[\vec{\gamma_k}]=\vec{0_{Q}}$ and $cov(\vec{\gamma_k}) = \mat{G} = diag(\sigma_1^2, \dots, \sigma_Q^2)$ where all entries are non-negative, $\vec{\epsilon_k} \in \mathds{R}^{N_k}$ is the vector of residual errors with $E[\vec{\epsilon_k}] = \vec{0_{N_k}}$ and $cov(\vec{\epsilon_k}) = \sigma_\epsilon^2 \mat{I_{N_k}}$. The random effects are independent of the residual errors. For $j = 1, \dots, Q$, if $\sigma_j^2>0$, then the effect of the corresponding predictor differs across studies, and if $\sigma_j^2=0$, then the predictor has the same effect in each study. Model \ref{model1} is nonparametric in the fixed effects but linear in the random effects. Similar models have been considered by \cite{wang1998mixed} and \cite{gu2005optimal}. The linear mixed effects model 
\begin{flalign}\label{model1b}
\vec{Y_{k}} = \mat{X_{k}}\vec{\beta} + \mat{Z_{k}}\vec{\gamma_k} + \vec{\epsilon_{k}}
\end{flalign}
where $\vec{\beta} \in \mathds{R}^{P}$ is the vector of fixed effects is a special case of Model \ref{model1}. 

The relationship between the predictors and the outcome in a given study can be seen as a perturbation of the population-level relationship described by $\mat f$. The degree of heterogeneity in predictor-outcome relationships across studies can be summarized by the sum of the variances of the random effects divided by the number of fixed effects: $\overline{\sigma^2} = tr(\mat{G}) / P$. We are interested in comparing the performance of two multi-study prediction approaches as $\overline{\sigma^2}$ varies: 1) merging all of the studies and fitting a single linear regression model, and 2) fitting a linear regression model on each study and forming a multi-study ensemble by taking a weighted average of the predictions from each study-specific model. These multi-study strategies can be applied to any type of study-specific model, including non-parametric machine learning models, but our theoretical analysis will be restricted to linear models, as they have a closed form solution. Comparisons involving machine learning models without a closed form solution are also interesting and will be explored in the simulations.

\subsection{Ridge Regression Prediction Models}

We will use the $K$ studies to train merged and ensemble versions of ridge regression prediction models, allowing for basis expansion of the predictors. This approach encompasses many commonly used linear models as special cases, including ordinary least squares and penalized spline regression. Though we focus on a class of linear prediction models, with an appropriate choice of bases these models can in theory approximate any sufficiently smooth function to any level of precision \citep{stone1948generalized}. 

Consider a pre-specified set of basis functions for each predictor. Suppose there are $M_p$ one-dimensional basis functions for predictor $p$, given by $\{ h_{1p}, \dots, h_{M_p p} \}$. Let $\vec h(\vec{X_{jk}}) = (h_{11}(X_{1jk}), \dots, h_{M_1 1}(X_{1jk}), \dots, h_{1P}(X_{Pjk}), \dots, h_{M_P P}(X_{Pjk}))$. Define the basis-expanded design matrix
\[\mat{\tilde{X}_{k}} = \begin{bmatrix} \vec{h}\left(\mat{X_{1k}}\right) | \dots | \vec{h}\left(\mat{X_{N_k k}} \right) \end{bmatrix}^T  \in \mathds{R}^{N_k \times M}\]
where $M = \sum_p M_p$. Let the first column of $\mat{\tilde{X}_{k}}$ be a column of 1's for the intercept if the prediction model will include an intercept. For the merged learner, we fit the model 
\begin{flalign}\label{model2a}
E[\vec{Y}] = \mat{\tilde{X}} \vec{\beta} 
\end{flalign}
where $\vec{Y} = (\vec{Y_1}, \dots, \vec{Y_K}) \in \mathds{R}^{\sum_{k=1}^K N_k}$ and $\mat{\tilde{X}} =  \begin{bmatrix} \mat{\tilde{X}_1}^T | \dots | \mat{\tilde{X}_K}^T\end{bmatrix}^T \in \mathds{R}^{\sum_{k=1}^K N_k \times M}$. The estimator for $\vec{\beta}$ is
\begin{flalign}\label{merged}
\vec{\hat{\beta}_{merge}} = (\mat{\tilde{X}}^T \mat{\tilde{X}} + \lambda \mat{I_{M}^-})^{-1}\mat{\tilde{X}}^T \vec{Y}
\end{flalign}
where $\lambda \geq 0$ is a pre-specified regularization parameter and $\mat{I_M^-}$ is defined as $diag(0, 1, \dots, 1)$ if there is an intercept and $\mat{I_M}$ otherwise.

For the ensemble learner, we first fit the models
\begin{flalign}
E[\vec{Y_{k}}] = \mat{\tilde{X}_{k}} \vec{\beta_k} 
\end{flalign}
separately in studies $k=1,\dots,K$ to obtain estimators
\begin{flalign}
\vec{\hat{\beta}_k} = (\mat{\tilde{X}_k}^T \mat{\tilde{X}_k} + \lambda_k \mat{I_{M}^-})^{-1}\mat{\tilde{X}_k}^T \vec{Y_k}.
\end{flalign}
Then the ensemble estimator, based on pre-specified weights $w_k$ satisfying $\sum\limits_{k=1}^K w_k = 1$, is 
\begin{flalign}\label{CSL}
\vec{\hat{\beta}_{ens}} = \sum_{k=1}^K w_k \vec{\hat{\beta}_{k}}.
\end{flalign}
Examples of basis functions for a given predictor include $h(x)=x$, which gives a model linear in the original predictor, $h_m(x)=x^m$ for $m=0,\dots,S$, which gives a degree $S$ polynomial regression model, and basis splines of degree $S$ with knots $\xi_1, \dots, \xi_T$:
\begin{flalign}\label{bsplines}
h_{i,0} &= I[\xi_i \leq x < \xi_{i+1}] &\\
h_{i,s}(x) &= \frac{x-\xi_i}{\xi_{i+s} - \xi_i} h_{i, s-1}(x) + \frac{\xi_{i+s+1}-x}{\xi_{i+s+1} - \xi_{i+1}} h_{i+1, s-1}(x) \tag{$s=1,\dots,S$} \notag
\end{flalign}
Setting $\lambda=0$ is equivalent to applying ordinary least squares with design matrix $\mat{\tilde{X}}$. In order to derive analytic results, we assume that the weights $w_k$ and the regularization parameters $\lambda$ and $\lambda_k$ are predetermined. While in some cases these parameters can be chosen using independent data \citep{tsybakov2014aggregation}, in practice it is preferable to estimate them from the training data. We study the effects of this estimation strategy in simulations. 

There are connections between ensembling, meta-analysis, and linear mixed effects models. For linear regression, averaging predictions across study-specific learners is equivalent to averaging the estimated coefficient vectors across study-specific learners and then computing predictions, so ensembling based on linear regression on scaled variables has points in common with meta-analysis of effect sizes. When $P=1$, a standard meta-analysis is also a weighted average of $\hat{\beta}_k$'s, though typically weights reflect precision of estimates rather than cross-study features of predictions. When $P>1$, $\vec{\hat{\beta}_{ens}}$ weights each dimension of the coefficient vector equally in a given study while meta-analytic approaches, which involve either performing separate univariate meta-analyses for each predictor or performing a multivariate meta-analysis (for example, see \cite{jackson2010extending, jackson2011multivariate}), do not impose this constraint. The fixed effect estimator for the linear mixed effects model when $P=1$ is a weighted average of the study-specific least squares estimators, which is a form of ensembling. In general, ensembling linear models with appropriate weights may provide a reasonable approximation of a mixed effects model, but ensembling is applicable to a broader range of scenarios than linear models. 

\subsection{Performance Metric} 

We compare the performance of the merged and ensemble ridge regression models on a new test dataset, not used for training, with fixed design matrix $\mat{X_0}$ and outcome vector $\vec{Y_0}$ (generated under Model \ref{model1}). The goal is to identify conditions under which ensembling has lower mean squared prediction error, conditional on $\mat{X_0}$, than merging, i.e.
\[ E[\| \mat {Y_0} - \tilde{\mat{X_0}} \vec{\hat\beta_{ens}}\|_2^2| \mat{{X}_0} ] \leq E[\| \mat {Y_0} - \tilde{\mat{X_0}} \vec{\hat\beta_{merge}} \|_2^2| \mat{{X}_0} ]\]
where expectations are taken with respect to $\vec{Y_0}|\mat{X_0}$. Also, given a vector $(u_1, \dots, u_m)$ of length~$m$, $\|(u_1, \dots, u_m)\|_2 = \sqrt{\sum_{i=1}^m u_i^2}$ is the L2 norm. Since we are evaluating the predictions of ridge regression models rather than linear mixed effects models and the test dataset is independent of the training datasets, the above inequality does not include any random effects. Moreover, predictions are utilizing $\mat{X_0}$ only, while $\vec{Y_0}$ is held as the gold standard. 

\section{Results}

\subsection{Overview}
We consider two cases for the structure of $\mat G$: equal variances and unequal variances. Let $\sigma_{(1)}^2, \dots, \sigma_{(D)}^2$ be the distinct values on the diagonal of $\mat G$ and let $J_d$ be the number of random effects with variance $\sigma_{(d)}^2$. In the equal variances case where $D=1$ and $\sigma_{j}^2 = \sigma^2$ for $j=1,\dots,Q$, Theorem 1 provides a necessary and sufficient condition for the ensemble learner to outperform the merged learner. In the unequal variances case, Theorem 2 provides sufficient conditions under which the ensemble learner outperforms the merged learner and vice versa. These conditions allow us to characterize a transition point in terms of the heterogeneity measure $\overline{\sigma^2}$ between a setting that favors merging and a setting that favors ensembling. We also provide optimal ensemble weights in Proposition \ref{prop1}, state special cases of the results for prediction models trained using ordinary least squares in Corollaries 1-3, provide an asymptotic version of the transition point for least squares in Corollary 4, and discuss how to interpret the results. For conciseness, we define the following notation.
\begin{flalign}
&\text{Let } \mat{\tilde{R}_k} = \mat{\tilde{X}_k}^T\mat{\tilde{X}_k} + \lambda_k \mat{I_{M}^-}. \\
&\text{Let } \mat{\tilde{R}} = \mat{\tilde{X}}^T\mat{\tilde{X}} + \lambda \mat{I_{M}^-}. \\
&\text{Let } \mat{\tilde{A}_k} = \mat{\tilde{X}_k} \mat{\tilde{R}_k}^{-1} \mat{\tilde{X}_0}^T\mat{\tilde{X}_0} \mat{\tilde{R}_k}^{-1} \mat{\tilde{X}_k}^T.\\
&\text{Let } \mat{\tilde{C}_k} = \mat{\tilde{X}_k} \mat{\tilde{R}}^{-1} \mat{\tilde{X}_0}^T \mat{\tilde{X}_0} \mat{\tilde{R}}^{-1}  \mat{\tilde{X}_k}^T.\\
&\text{Let } \vec{b_{ens}} = Bias(\mat{\tilde{X}_0} \vec{\hat{\beta}_{ens}} ) = \sum_{k=1}^K w_k \mat{\tilde{X}_0} \mat{\tilde{R}_k}^{-1} \mat{\tilde{X}_k}^T  \vec f(\mat{X_k}) - \vec f(\mat{X_0}) \\
&\text{be the bias of the ensemble predictions in the test set.}\nonumber \\
&\text{Let } \vec{b_{merge}} = Bias(\mat{\tilde{X}_0} \vec{\hat{\beta}_{merge}} )  = \mat{\tilde{X}_0} \mat{\tilde{R}}^{-1} \mat{\tilde{X}}^T \vec f(\mat{X}) - \vec f(\mat{X_0}) \\
&\text{ be the bias of the merged predictions in the test set.} \nonumber
\end{flalign}

\subsection{Main Results: Ridge Regression}\label{main}

\begin{theorem}\label{thm1}
Suppose $\sigma_{j}^2 = \sigma^2$ for $j=1,\dots,Q$ and 
\begin{flalign}
tr \left\{ \sum\limits_{k=1}^K \mat{Z_k}^T \left( \mat{\tilde{C}_k} - w_k^2 \mat{\tilde{A}_k} \right) \mat{Z_k}   \right\} > 0.  \label{cond:cond1}
\end{flalign}
Define
{\begin{flalign}
\tau = \frac{Q}{P} \times \frac{ \sigma_\epsilon^2 tr \left\{ \sum\limits_{k=1}^K  \left(w_k^2\mat{\tilde{A}_k} - \mat{\tilde{C}_k} \right)  \right\} + \vec{b_{ens}}^T \vec{b_{ens}} - \vec{b_{merge}}^T \vec{b_{merge}}}      {tr \left\{ \sum\limits_{k=1}^K \mat{Z_k}^T \left( \mat{\tilde{C}_k} - w_k^2 \mat{\tilde{A}_k} \right) \mat{Z_k}   \right\}}.
\end{flalign}}
Then $E[\| \mat {Y_0} - \mat{\tilde{X}_0} \vec{\hat\beta_{ens}} \|_2^2 | \mat{{X}_0}] \leq E[\| \mat {Y_0} - \mat{\tilde{X}_0} \vec{\hat\beta_{merge}} \|_2^2 | \mat{{X}_0}]$ if and only if $\overline{\sigma^2} \geq {\tau}$.
\end{theorem}

Theorem \ref{thm1} is a special case of Theorem \ref{thm2} below, which is proved in Appendix A (Supplementary Materials). By Theorem \ref{thm1}, for any fixed set of weights that satisfy Condition \ref{cond:cond1}, $\tau$ represents a transition point between a setting where merging outperforms ensembling and a setting where ensembling outperforms merging. The transition point depends on the true population-level relationship $\vec f$ between the predictors and the outcome since $\vec f$ is part of the bias terms $\vec{b_{ens}}$ and $\vec{b_{merge}}$, so we need to estimate $\vec f$ in order to estimate the transition point. Sensitivity to misspecification of $\vec f$ will be studied in simulations in Section \ref{simresults}.

\begin{theorem}\label{thm2}
\begin{enumerate}
    \item Suppose
\begin{flalign}\label{cond2a}
\max\limits_d \sum\limits_{j:\sigma_j^2=\sigma_{(d)}^2} \left\{\sum\limits_{k=1}^K  \mat{Z_k}^T \left( \mat{\tilde{C}_k} - w_k^2 \mat{\tilde{A}_k} \right) \mat{Z_k}    \right\}_{jj} > 0
\end{flalign}
and define
{\begin{flalign}
{\tau}_{1} = \frac{ \sigma_\epsilon^2 tr \left\{ \sum\limits_{k=1}^K  \left(w_k^2\mat{\tilde{A}_k} - \mat{\tilde{C}_k} \right)  \right\} + \vec{b_{ens}}^T \vec{b_{ens}} - \vec{b_{merge}}^T \vec{b_{merge}}}    {P \max\limits_d \frac{1}{J_d} {\sum\limits_{j:\sigma_j^2=\sigma_{(d)}^2} \left\{\sum\limits_{k=1}^K  \mat{Z_k}^T \left( \mat{\tilde{C}_k} - w_k^2 \mat{\tilde{A}_k} \right) \mat{Z_k}    \right\}_{jj}}}.
\end{flalign}}
Then $E[\| \mat {Y_0} - \mat{\tilde{X}_0} \vec{\hat\beta_{ens}} \|_2^2 | \mat{{X}_0}] \geq E[\| \mat {Y_0} - \mat{\tilde{X}_0} \vec{\hat\beta_{merge}} \|_2^2 | \mat{{X}_0}]$ when $\overline{\sigma^2} \leq {\tau}_{1}$.
\item Suppose \begin{flalign}\label{cond2b}
\min\limits_d \sum\limits_{j:\sigma_j^2=\sigma_{(d)}^2} \left\{\sum\limits_{k=1}^K  \mat{Z_k}^T \left( \mat{\tilde{C}_k} - w_k^2 \mat{\tilde{A}_k} \right) \mat{Z_k}    \right\}_{jj} > 0
\end{flalign}
and define
{\begin{flalign}{\tau}_{2} = \frac{ \sigma_\epsilon^2 tr \left\{ \sum\limits_{k=1}^K  \left(w_k^2\mat{\tilde{A}_k} - \mat{\tilde{C}_k} \right)  \right\} + \vec{b_{ens}}^T \vec{b_{ens}} - \vec{b_{merge}}^T \vec{b_{merge}}}    {P \min\limits_d \frac{1}{J_d} {\sum\limits_{j:\sigma_j^2=\sigma_{(d)}^2} \left\{\sum\limits_{k=1}^K  \mat{Z_k}^T \left( \mat{\tilde{C}_k} - w_k^2 \mat{\tilde{A}_k} \right) \mat{Z_k}    \right\}_{jj}}} .
\end{flalign}}
Then
$E[\| \mat {Y_0} - \mat{\tilde{X}_0} \vec{\hat\beta_{ens}} \|_2^2 | \mat{{X}_0}] \leq E[\| \mat {Y_0} - \mat{\tilde{X}_0} \vec{\hat\beta_{merge}} \|_2^2 | \mat{{X}_0}]$ when
$\overline{\sigma^2} \geq {\tau}_{2}$.
\end{enumerate}
\end{theorem}

A proof of Theorem 2 is provided in Appendix A (Supplementary Materials). Theorem 2 states that in a more general scenario where the random effects do not necessarily have the same variance, there is a transition interval such that the merged learner outperforms the ensemble learner when $\overline{\sigma^2}$ is smaller than the lower bound of the interval and the ensemble learner outperforms the merged learner when $\overline{\sigma^2}$ is greater than the upper bound of the interval. While Theorem 2 does not provide information on the relative performance of the learners within the interval, we will investigate this in the simulations.

\begin{proposition}\label{prop1}
Let $v_k = tr(\mat{G} \mat{Z_k}^T \mat{\tilde{C}_k} \mat{Z_k}  + \sigma_\epsilon^2 \mat{\tilde{A}_k} )$ and $\mat{b_k} = \mat{\tilde{X}_0} \mat{\tilde{R}_k}^{-1} \mat{\tilde{X}_k}^T  \vec f(\mat{X_k}) - \vec f(\mat{X_0})$.
Then the optimal weights for the ridge regression ensemble are
\begin{flalign}
\vec{w_k} = \frac{\sum_{j=1}^K (\mat E^{-1})_{kj}}{\sum_{i=1}^K \sum_{j=1}^K (\mat E^{-1})_{kj}}
\end{flalign}
where $\mat E \in \mathds{R}^{K \times K}$ has entries $(\mat E)_{kk} = v_k+ \mat{b_k}^T \mat{b_k} $ and $(\mat E)_{jk} =  \mat{b_j}^T \mat{b_k}$ for $j \neq k$.
\end{proposition}

In general, the optimal weights depend on the true population-level relationship between the predictors and the outcome, $\vec f$, via the bias terms $\vec{b_k}$. The optimal weights also depend on $\mat{G}$ and $\sigma_\epsilon^2$. Using a first-order Taylor approximation, the optimal $\vec{w_k}$ is approximately proportional to $(v_k + \vec{b_k}^T\vec{b_k})^{-1} - (v_k + \vec{b_k}^T\vec{b_k})^{-2} - \sum_{j \neq k} \vec{b_j}^T\vec{b_k} (v_j + \vec{b_j}^T\vec{b_j})^{-1} (v_k + \vec{b_k}^T\vec{b_k})^{-1}$,
where the first two terms depend on the inverse mean squared prediction error and the third term depends on the covariances between the bias terms from study $k$ and the bias terms from other studies, as well as the magnitudes of the studies' prediction errors. If the magnitudes of the prediction error and bias term for study $k$ are held fixed and the other studies are held fixed, then the optimal weight for study $k$ increases as it becomes less correlated with the other studies. This is similar to a result from \cite{krogh1995neural} that decomposes the prediction error of an ensemble into a term that depends on the prediction errors of the individual learners and a correlation term that quantifies the disagreement across the individual learners; the decomposition implies that if the individual prediction errors are held fixed, then the performance of the ensemble improves as the correlation term decreases.

The transition point under any fixed weighting scheme provides an upper bound for the transition point under optimal weights, which can be calculated numerically. In practice, $\vec f$, $\mat G$, and $\sigma_\epsilon^2$ are not known, so the optimal weights need to be estimated. Weight estimation increases the variance of the ensemble learner and may result in a higher transition point than when the optimal weights are known because the numerator of the transition point includes the variance of the ensemble learner. However, as seen in Supplementary Figure A.8, our simulations suggest that if $\vec f$, $\mat G$, and $\sigma_\epsilon^2$ can be reasonably estimated - for example, via a mixed effects model - then the estimation will have little impact on the transition point.

\subsection{Special Case: Least Squares}\label{leastsquares}

Least squares is a special case of ridge regression where the regularization parameter is 0. The lack of regularization allows for a simplification of the results from Section \ref{main}. In particular, when the generating model is Model \ref{model1b}, which is linear in $\mat{X_k}$, a least squares model based on the original predictors $\mat{X_k}$ is correctly specified and unbiased. Here, we present Corollaries 1-3, which are special cases of Theorem 1, Theorem 2, and Proposition 1, as well as Corollary 4, which provides an asymptotic version of Theorem 1 as the number of training studies tends to infinity. For the results below, we assume $\mat{\tilde{X}_k} = \mat{X_k}$. In Theorems 1 and 2, we used $\mat{\tilde{A}_k}$ and $\mat{\tilde{C}_k}$ in the analytic expressions for conciseness, but below we present the results in terms of the original predictor matrices.

\begin{corollary}\label{cor1}
Suppose $\sigma_{j}^2 = \sigma^2$ for $j=1,\dots,Q$, $\vec f$ is a linear function of $\mat{X_k}$, $\lambda=\lambda_k=0$, and
\begin{flalign}
tr \left[ \sum\limits_{k=1}^K  \left\{ \mat{Z_k}^T  \mat{{X}_k} (\mat{X}^T \mat{X})^{-1}  \mat{X_0}^T \mat{X_0} (\mat{X}^T \mat{X})^{-1}  \mat{{X}_k}^T \mat{Z_k} - w_k^2  \mat{Z_0}^T \mat{Z_0}\right\} \right] > 0.  \label{cond:cond1b}
\end{flalign}
Define
{\begin{flalign}
\tau_{LS} = \frac{Q}{P} \times \frac{ \sigma_\epsilon^2  tr \left[ \left\{ \sum\limits_{k=1}^K w_k^2 (\mat{X_k}^T \mat{X_k})^{-1} - (\mat{X}^T \mat{X})^{-1} \right\} \mat{X_0}^T \mat{X_0}  \right] }      {tr \left[ \sum\limits_{k=1}^K  \left\{ \mat{Z_k}^T  \mat{{X}_k} (\mat{X}^T \mat{X})^{-1}  \mat{X_0}^T \mat{X_0} (\mat{X}^T \mat{X})^{-1}  \mat{{X}_k}^T \mat{Z_k} - w_k^2  \mat{Z_0}^T \mat{Z_0}\right\} \right]}
\end{flalign}}
Then $E[\| \mat {Y_0} - \mat{{X}_0} \vec{\hat\beta_{ens}} \|_2^2 | \mat{{X}_0}] \leq E[\| \mat {Y_0} - \mat{{X}_0} \vec{\hat\beta_{merge}} \|_2^2 | \mat{{X}_0}]$ if and only if $\overline{\sigma^2} \geq {\tau_{LS}}$.
\end{corollary}

Assuming $\mat{R_k}$ is not identical for all $k$, if equal weights $w_k=1/K$ are used, then Condition \ref{cond:cond1b} is satisfied due to Jensen's operator inequality \citep{hansen2003jensen} and $\tau_{LS}>0$, so the transition point always exists.

\begin{corollary}
Suppose $\vec f$ is a linear function of $\mat{{X}_k}$ and $\lambda=\lambda_k=0$.
\begin{enumerate}
    \item Suppose
\begin{flalign}
\max\limits_d \sum\limits_{j:\sigma_j^2=\sigma_{(d)}^2} \left[ \sum\limits_{k=1}^K  \left\{ \mat{Z_k}^T  \mat{{X}_k} (\mat{X}^T \mat{X})^{-1}  \mat{X_0}^T \mat{X_0} (\mat{X}^T \mat{X})^{-1}  \mat{{X}_k}^T \mat{Z_k} - w_k^2  \mat{Z_0}^T \mat{Z_0}\right\} \right]_{jj} > 0
\end{flalign}
and define
{\begin{flalign}
{\tau}_{LS,1} = \frac{\sigma_\epsilon^2  tr \left[ \mat{X_0}^T \mat{X_0} \left\{ \sum\limits_{k=1}^K w_k^2 (\mat{X_k}^T \mat{X_k})^{-1} - (\mat{X}^T \mat{X})^{-1} \right\} \right]}   {P \max\limits_{d} \frac{1}{J_d} {\sum\limits_{j:\sigma_j^2=\sigma_{(d)}^2} \left[ \sum\limits_{k=1}^K  \left\{ \mat{Z_k}^T  \mat{{X}_k} (\mat{X}^T \mat{X})^{-1}  \mat{X_0}^T \mat{X_0} (\mat{X}^T \mat{X})^{-1}  \mat{{X}_k}^T \mat{Z_k} - w_k^2  \mat{Z_0}^T \mat{Z_0}\right\} \right]_{jj}}} .
\end{flalign}}
Then $E[\| \mat {Y_0} - \mat{{X}_0} \vec{\hat\beta_{ens}} \|_2^2 | \mat{{X}_0}] \geq E[\| \mat {Y_0} - \mat{{X}_0} \vec{\hat\beta_{merge}} \|_2^2 | \mat{{X}_0}]$ when $\overline{\sigma^2} \leq {\tau}_{LS,1}$.
\item Suppose \begin{flalign}
\min\limits_d \sum\limits_{j:\sigma_j^2=\sigma_{(d)}^2} \left[ \sum\limits_{k=1}^K  \left\{ \mat{Z_k}^T  \mat{{X}_k} (\mat{X}^T \mat{X})^{-1}  \mat{X_0}^T \mat{X_0} (\mat{X}^T \mat{X})^{-1}  \mat{{X}_k}^T \mat{Z_k} - w_k^2  \mat{Z_0}^T \mat{Z_0}\right\} \right]_{jj} > 0
\end{flalign}
and define
{\begin{flalign}{\tau}_{LS,2} = \frac{\sigma_\epsilon^2  tr \left[ \mat{X_0}^T \mat{X_0} \left\{ \sum\limits_{k=1}^K w_k^2 (\mat{X_k}^T \mat{X_k})^{-1} - (\mat{X}^T \mat{X})^{-1} \right\} \right] }   {P \min\limits_{d} \frac{1}{J_d} {\sum\limits_{j:\sigma_j^2=\sigma_{(d)}^2} \left[ \sum\limits_{k=1}^K  \left\{ \mat{Z_k}^T  \mat{{X}_k} (\mat{X}^T \mat{X})^{-1}  \mat{X_0}^T \mat{X_0} (\mat{X}^T \mat{X})^{-1}  \mat{{X}_k}^T \mat{Z_k} - w_k^2  \mat{Z_0}^T \mat{Z_0}\right\} \right]_{jj}}} .
\end{flalign}}
Then $E[\| \mat {Y_0} - \mat{{X}_0} \vec{\hat\beta_{ens}} \|_2^2 | \mat{{X}_0}] \leq E[\| \mat {Y_0} - \mat{{X}_0} \vec{\hat\beta_{merge}} \|_2^2 | \mat{{X}_0}]$ when $\overline{\sigma^2} \geq {\tau}_{LS,2}$.
\end{enumerate}
\end{corollary}

\begin{corollary}\label{cor2}
Suppose $\vec f$ is a linear function of $\mat{{X}_k}$ and $\lambda_k=0$. Then the optimal weights for the least squares ensemble are
\begin{flalign}
w_k = \dfrac{\{tr(\mat{G} \mat{Z_0}^T \mat{Z_0}) + \sigma_\epsilon^2 tr( (\mat{X_k}^T \mat{X_k})^{-1}  \mat{X_0}^T \mat{X_0} )\}^{-1}}{\sum_{k=1}^K \{tr(\mat{G} \mat{Z_0}^T \mat{Z_0}) + \sigma_\epsilon^2 tr((\mat{X_k}^T \mat{X_k})^{-1} \mat{X_0}^T \mat{X_0} )\}^{-1}}.
\end{flalign}
\end{corollary}

In this setting, the optimal weight for study $k$ is proportional to the inverse mean squared prediction error of the least squares learner trained on that study.

\begin{corollary}\label{cor4}
Suppose $\vec f$ is a linear function of $\mat{{X}_k}$ and $\lambda=\lambda_k=0$, and there exist positive definite matrices $\mat{M_1}, \mat{M_2}, \mat{M_3} \in \mathds{R}^{P \times P}$ such that as $K \to \infty$,
\begin{enumerate}
\item $\frac{1}{K} \sum\limits_{k=1}^K \mat{X_k}^T \mat{X_k} \to \mat{M_1}$ 
\item $\frac{1}{K} \sum\limits_{k=1}^K (\mat{X_k}^T \mat{X_k})^{-1} \to \mat{M_2}$ 
\item $\frac{1}{K} \sum\limits_{k=1}^K \mat{X_k}^T  \mat{Z_k} \mat{Z_k}^T \mat{X_k} \to \mat{M_3}$
\end{enumerate}
where $\to$ denotes almost sure convergence. Let $w_k = 1/K$. 
\begin{enumerate}
\item If \begin{flalign}
\max\limits_d \sum\limits_{j:\sigma_j^2=\sigma_{(d)}^2}  \left( \mat{M_1}^{-1}  \mat{M_3} \mat{M_1}^{-1} \mat{X_0}^T \mat{X_0} - \mat{Z_0}^T \mat{Z_0} \right)_{jj}  > 0
\end{flalign}
then
\begin{flalign} \tau_{LS,1} \to \frac{\sigma_\epsilon^2}{P} \times \frac{tr(\mat{M_2} \mat{X_0}^T \mat{X_0}) - tr(\mat{M_1}^{-1} \mat{X_0}^T \mat{X_0})}{ \max\limits_{d} \frac{1}{J_d} \sum\limits_{j:\sigma_j^2=\sigma_{(d)}^2}  \left( \mat{M_1}^{-1}  \mat{M_3} \mat{M_1}^{-1} \mat{X_0}^T \mat{X_0} - \mat{Z_0}^T \mat{Z_0} \right)_{jj}}
\end{flalign}
\item If \begin{flalign}
\min\limits_d \sum\limits_{j:\sigma_j^2=\sigma_{(d)}^2}  \left( \mat{M_1}^{-1}  \mat{M_3} \mat{M_1}^{-1} \mat{X_0}^T \mat{X_0} - \mat{Z_0}^T \mat{Z_0} \right)_{jj}  > 0
\end{flalign} then
\begin{flalign} 
\tau_{LS,2} \to \frac{\sigma_\epsilon^2}{P} \times \frac{tr(\mat{M_2} \mat{X_0}^T \mat{X_0}) - tr(\mat{M_1}^{-1} \mat{X_0}^T \mat{X_0})}{ \min\limits_{d} \frac{1}{J_d} \sum\limits_{j:\sigma_j^2=\sigma_{(d)}^2}  \left( \mat{M_1}^{-1}  \mat{M_3} \mat{M_1}^{-1} \mat{X_0}^T \mat{X_0} - \mat{Z_0}^T \mat{Z_0} \right)_{jj} }
\end{flalign}
\end{enumerate}

\end{corollary}

Corollary \ref{cor4} presents an asymptotic version of Corollary \ref{cor2} as the number of studies goes to infinity. If all study sizes are equal to $n$ and the predictors are independent and identically distributed within and across studies, then $\mat{M_1} = E[\mat{X_k}^T \mat{X_k}]$, $\mat{M_2} = E[(\mat{X_k}^T \mat{X_k})^{-1}]$, and $\mat{M_3} = E[\mat{X_k}^T  \mat{Z_k} \mat{Z_k}^T \mat{X_k} ]$. In the special case where $P=Q=1$ and the predictor follows $N(0, v)$, $\tau_{LS,1}=\tau_{LS,2}$ converges to
\begin{flalign} \label{cor1_limit}
\frac{\sigma_\epsilon^2}{(n-2)v}, \end{flalign}
so asymptotically the transition point is controlled simply by the variance of the residuals, the variance of the predictor, and the study sample size.

\subsection{Interpretation}\label{interp}

The covariance matrices of linear regression coefficient estimators can be written as a sum of two components, one driven by between-study variability (the random effects in Model \ref{model1}) and one driven by within-study variability (the residual errors in Model \ref{model1}). For example, when all predictors have random effects, i.e. $\mat{Z_k} = \mat{X_k}$, the covariance of the least squares ensemble is
\[ cov(\vec{\hat \beta_{LS, ens}}) = \sum_{k=1}^K w_k^2 \mat{G} + \sigma_\epsilon^2 \sum_{k=1}^K w_k^2 (\mat{X_k}^T \mat{X_k})^{-1} \]
and the covariance of the merged least squares learner is
\[cov(\vec{\hat \beta_{LS, merge}}) = (\mat{X}^T \mat{X})^{-1} \sum_{k=1}^K[\mat{X_k}^T \mat{X_k} \mat{G} \mat{X_k}^T  \mat{X_k}](\mat{X}^T \mat{X})^{-1} + \sigma_\epsilon^2 (\mat{X}^T \mat{X})^{-1}.\]
Since the merged learner ignores between-study heterogeneity, the trace of its first component is generally larger than that of the ensemble learner. However, since the merged learner is trained on a larger sample, the trace of its second component is generally smaller than that of the ensemble learner. When merged and ensemble learners are trained using correctly specified least squares models, they are unbiased, so the transition point depends on the trade-off between these two components. When $P=1$, Expression \ref{cor1_limit} shows that having a higher-variance predictor favors ensembling over merging, since increasing the variance of the predictor amplifies the impact of the random effect.

Unlike least squares estimators, ridge regression estimators are biased as a result of regularization even when the models are correctly specified. The transition point for ridge regression depends on the regularization parameters used on the merged and individual datasets. It also depends on the true relationship $\vec f$ between the predictors and the outcome through the squared bias terms in the mean squared prediction errors of the merged and ensemble learners, so an estimate of $\vec f$ is needed to compute the expressions in Theorems \ref{thm1} and \ref{thm2}. These expressions can vary considerably for different choices of regularization parameters and different forms of $\vec f$. We did not provide the asymptotic results for ridge regression as $K \to \infty$ with the study sizes held constant because this scenario is not entirely fair to the ensemble learner. For $P > N_k$ and sufficiently large $K$, the merged learner will be in a low-dimensional setting where the number of samples exceeds the number of predictors, while the ensemble learner will remain in the high-dimensional setting. As $K \to \infty$, the bias term approaches 0 for the merged learner, assuming $\lambda/K \to 0$, but not for the ensemble learner, which suggests that when $K$ is sufficiently large, merging will always yield lower mean squared prediction error than ensembling. 

In general, the transition points for least squares and ridge regression depend on the design matrix of the test set. However, the test design matrix drops out when it is a scalar multiple of an orthogonal matrix. For example, this occurs when $P=1$.

\section{Simulations}\label{sim}

We conducted simulations to verify the theoretical results for least squares and ridge regression and to compare them to the empirical transition points for three methods for which we cannot derive a closed-form solution: lasso, single hidden layer neural network, and random forest. We also ran a linear mixed effects model. We used the R packages {\tt glmnet, nnet, randomForest} and {\tt nlme} for ridge/lasso, neural networks, random forests, and linear mixed effects models. 

\subsection{Data Generation}

The simulations were based on gut microbiome data from studies in the {\tt curatedMetagenomicData} R package, which was also the data source for our data example. In particular, we used marker abundance measurements from {\tt curatedMetagenomicData} studies as predictors in the simulations. We used a generating model with $P=10$ predictors, out of which $Q=5$ have random effects. The outcome for individual $j$ in study $k$ is generated by the model
\begin{flalign}\label{model1_sim}
Y_{jk} = f_0(\vec{X_{jk}}) + \vec{\gamma_k}^T \vec{Z_{jk}} + \epsilon_{jk}
\end{flalign}
where $\vec{\gamma_k}$ follows $MVN(\vec 0, \mat G)$ with $\mat G = diag(\sigma_1^2, \dots, \sigma_Q^2)$ for varying values of $\sigma_1^2, \dots, \sigma_Q^2$, $\epsilon_{jk}$ follows $N(0, \sigma_\epsilon^2)$ with $\sigma_\epsilon^2 = 1$,
{\begin{flalign}\label{f_sim}
f_0(X_{jk}) &= 0.61 h_{1,0}(X_{1jk}) + 0.18 h_{1,1}(X_{1jk})+ 0.20 h_{1,2}(X_{1jk}) + 0.06 h_{1,3}(X_{1jk}) - \\
& \phantom{==} 0.07 X_{2jk} + 0.47 X_{3jk} + 0.02 X_{4jk} + 0.04 X_{5jk} +0.51 X_{6jk} + 0.14 X_{7jk} - \notag \\
& \phantom{==} 0.38 X_{8jk} - 0.21 X_{9jk} - 0.13 X_{10jk}, \notag
\end{flalign}}
where $h_{1,0}, \dots, h_{1,3}$ are cubic basis splines as defined in Equation \ref{bsplines} with a single knot at 0, the coefficients were randomly generated from $N(0, 0.3)$, and $\vec{Z_jk}=(X_{3jk},X_{4jk},X_{5jk},X_{6jk},X_{7jk})$.

We used $K=5$ datasets of size $N_k=50$ to train the models and evaluated the mean squared prediction error in a test dataset, $k=0$, of size 100. For each study $k$, individuals were sampled from a separate randomly selected dataset in the {\tt curatedMegatenomicData} R package and outcomes were simulated under Model \ref{model1_sim} conditional on the real predictor values.

Our main simulation analysis was based on the assumptions of Theorem \ref{thm1}, where the random effects have equal variances $\sigma_1^2 = \dots = \sigma_Q^2 = \sigma^2$. We considered choices of $\mat G$ corresponding to different levels of heterogeneity ranging from $\overline{\sigma^2} = \sigma^2/2 = 0$ to $\overline{\sigma^2} = \sigma^2/2 = 3 \tau_R$, where $\tau_R$ is the theoretical transition point for ridge regression from Theorem \ref{thm1}. We performed another analysis based on the assumptions of Theorem \ref{thm2}, where the random effects have unequal variances. For this analysis, we considered choices of $\mat G$ corresponding to levels of heterogeneity ranging from $\overline{\sigma^2} = 0$  (wherein the predictors have the same effects in each study) to $\overline{\sigma^2} = 2 \tau_{R,2}$, where $\tau_{R,2}$ is the theoretical upper bound of the transition interval from Theorem \ref{thm2}. For each choice of $\mat G$ and each study $k$, including the test study $k$=0, the outcomes were generated from $\mat{X_k}$ using the following steps: 1) sample the $Q$ random effects $\vec{\gamma_k}$ from $MVN(\vec 0, \mat G)$, 2) sample the residual errors $\vec{\epsilon_k}$ independently from $N(0, \sigma_\epsilon^2)$ with $\sigma_\epsilon^2 = 1$, and 3) generate $\vec{Y_k}$ using Model \ref{model1} and the functional form given in Equation \ref{f_sim}. 

In addition to the simulations with $K=5$ studies and diagonal $\mat G$, we performed sensitivity analyses considering scenarios with misspecified least squares/ridge regression/lasso models. For these we used the original predictors without basis expansion, $K=10$ and $K=15$ instead of $K=5$. Studies of size $N_k=100$ instead of $N_k=50$, and scenarios include correlated random effects. We also performed simulations investigating the effects of estimating the optimal ensemble weights and using 10-fold cross-validation to choose the ridge regression hyperparameters instead of pre-specifying them.

\subsection{Prediction Models}

We trained and tested the following approaches: a correctly specified linear mixed effects model and merged and ensemble learners based on least squares, ridge regression, lasso, neural networks, and random forests. For ridge regression, lasso, random forest, and neural network, the following hyperparameters were chosen using 5-fold cross-validation in a tuning dataset of size 250 that was not used for training: the regularization parameter for ridge regression and lasso, the number of predictors sampled at each split for random forest, and the number of nodes and weight decay parameter for neural network. We restricted to neural networks with a single hidden layer to limit the computational burden. The outcomes in the tuning dataset were generated under $\sigma^2=0$. The hyperparameters for the merged versions of the models were selected using the entire tuning dataset while the hyperparameters for the study-specific models were selected using a subset of size 50.

\subsection{Results}\label{simresults}

\begin{figure}[!htb]
\centering
\includegraphics[width=0.82\columnwidth]{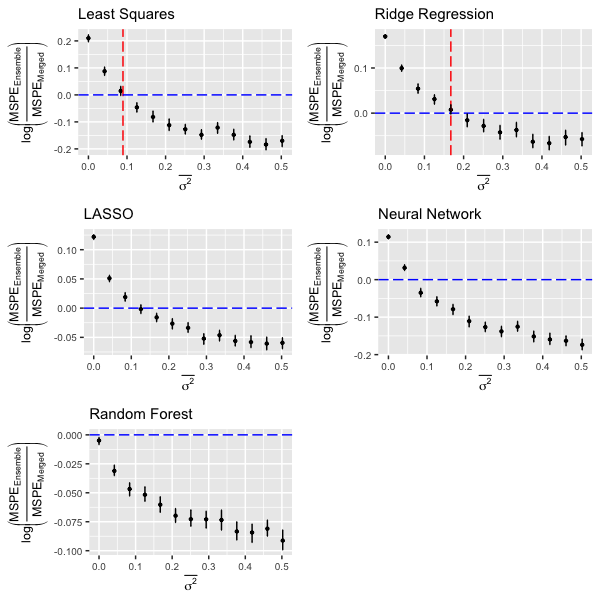}
\caption{Relative performance of multi-study ensembling and merging as a function of heterogeneity in the main simulation scenario where $K=5$, $N_k=50$, $P=10$, $Q=5$, and the random effects have equal variances. MSPE: mean squared prediction error. The vertical dashed lines correspond to the theoretical transition points calculated using Theorem \ref{thm1}. The empirical transition point occurs at the value of $\overline{\sigma^2}$ where the log ratio of the prediction errors for ensembling and merging is 0.}
\label{figure:thm1}
\end{figure}

\begin{figure}[htb]
\centering
\includegraphics[width=0.7\columnwidth]{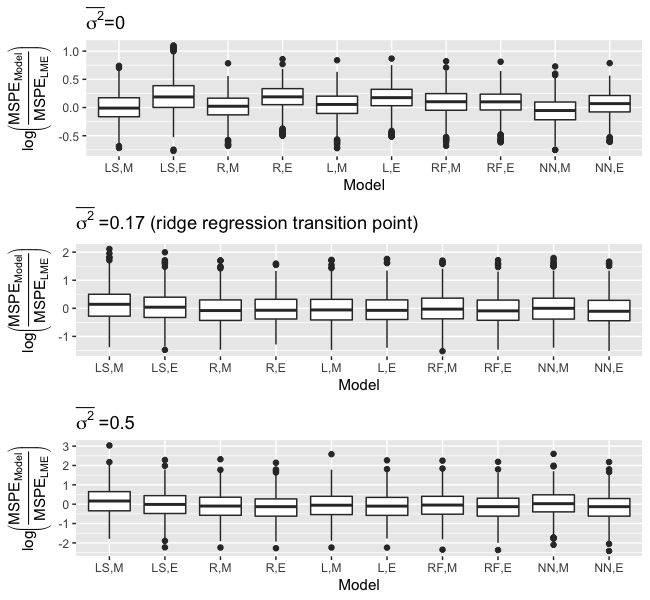}
\caption{Performance comparisons for three values of $\overline{\sigma^2}$ in main simulation scenario where $K=5$, $N_k=50$, $P=10$, $Q=5$, and the random effects have equal variances. MSPE: mean squared prediction error; LME: linear mixed effects model; LS,M: merged least squares learner; LS,E: ensemble learner based on least squares; R,M: merged ridge regression learner; R,E: ensemble learner based on ridge regression; L,M: merged lasso learner; L,E: ensemble learner based on lasso; NN,M: merged neural network; NN,E: ensemble learner based on neural networks; RF,M: merged random forest; RF,E: random forest.}
\label{figure:thm1_boxplots}
\end{figure}

\textbf{Transition Point Under Theoretical Assumptions.} For the main simulation analysis based on Theorem \ref{thm1}, Figure \ref{figure:thm1} shows the empirical transition points of the models and Figure \ref{figure:thm1_boxplots} shows the performance of the models relative to the data generating linear mixed effects model for three values of $\overline{\sigma^2}$. The empirical transition points under the other scenarios for which we derived theoretical results, i.e. assuming uncorrelated random effects, are shown in Supplementary Figures A.1-A.5. 

Figure \ref{figure:thm1_boxplots} shows that when $\overline{\sigma^2}=0$, the merged regression learners perform as well as or slightly better than the generating model and outperform the ensembles. In particular, the merged neural network performs better than the generating model. At the ridge regression transition point, all models perform similarly. Beyond the transition point, the models continue to perform similarly, with the ensembles slightly outperforming the merged learners. 

In all scenarios (Figures \ref{figure:thm1} and Supplementary Figures A.1-A.5), the empirical transition points for least squares and ridge regression agree with the theoretical results regardless of the number of training studies and whether the models were correctly specified. Also, in each scenario, the transition points of the different methods did not vary substantially except that the random forest either had no transition point because ensembling always outperformed merging (Figure \ref{figure:thm1} and Supplementary Figure A.2) or had a much earlier transition point than the other methods (Supplementary Figures A.3-A.5). This may be related to the fact that a single random forest is itself an ensemble. The empirical and theoretical performance of multi-study ensemble random forests has been studied in detail in \cite{Ramchandran:2019ga}, \cite{ramchandran2021ensembling}, and \cite{Ramchandran2021a}. In particular, \cite{ramchandran2021ensembling} derived theoretical results showing that ensembling random forests is expected to perform better than merging regardless of the number of studies. In some cases, it is also possible to improve on random forests by creating artificial studies via clustering and using a multi-study architecture on the results \citep{Ramchandran2021a}. 

\textbf{Robustness to Correlations between Random Effects.} In the scenarios with correlated random effects (Supplementary Figures A.6 and A.7), the least squares transition point occurred earlier than when the random effects were uncorrelated. In the positive correlation scenario, the ridge regression transition point also occurred earlier than when the random effects were uncorrelated, but the opposite was true in the negative correlation scenario. 
 
\textbf{Estimation of Optimal Weights and Tuning Hyperparameters via Cross-Validation.} As discussed in Section \ref{main}, the transition point is relatively robust to the usage of estimated optimal weights instead of fixed weights (Supplementary Figure A.8). However, when 10-fold cross-validation is used to tune the ridge regression hyperparameters, the concordance between the theoretical results and empirical performance depends on the level of heterogeneity. The theoretical transition point calculated by treating the cross-validated hyperparameters as fixed (i.e. ignoring the variability in the cross-validation procedure) was more accurate at predicting whether merging would outperform ensembling when there was low ($\overline{\sigma^2}<0.2$) or high ($\overline{\sigma^2}>0.5$) heterogeneity than when there was moderate heterogeneity (Supplementary Table A.1).

\section{Metagenomics Application}

Growing research on associations between the gut microbiome and health-related outcomes has motivated the development of prediction models based on metagenomic sequencing data \citep{fu2015gut, gupta2020predictive}. Predicting health outcomes from microbiome samples can help guide the design of controlled intervention studies that target the microbiome, for example using diet or probiotics. In our data example, we focus on cholesterol, a strong risk factor for cardiovascular disease. Studies have shown that the microbiome is associated with blood cholesterol levels \citep{fu2015gut, le2019intestinal, kenny2020cholesterol}. In particular, the metabolism of intestinal cholesterol by gut bacteria may reduce the amount of cholesterol absorbed from the intestine, resulting in lower blood cholesterol \citep{kenny2020cholesterol}. Recent work has identified bacterial genes involved in cholesterol metabolism \citep{kenny2020cholesterol}.

To illustrate in a practical example, we compared the performance of merging and ensembling in predict cholesterol levels from metagenomic sequencing data. We used datasets from the {\tt curatedMetagenomicData} R package \citep{pasolli2017accessible} (version 1.12.3), which contains a collection of curated, uniformly processed whole-metagenome sequencing data from multiple studies. In addition to metagenomic data such as gene marker abundance, it also includes demographic and clinical data. Cholesterol measurements were available for three of the studies that sequenced gut microbial DNA: 1) a study of Chinese type 2 diabetes patients and non-diabetic controls
\citep{qin2012metagenome}, 2) a study of middle-aged European women with normal, impaired or diabetic glucose control \citep{karlsson2013gut}, and 3) a study of patients with a family history of type 1 diabetes \citep{heintz2017integrated}. We used samples from female patients in these studies. The sample sizes were 151, 145, and 32 for the three studies.

We used merging and ensembling to train least squares, ridge regression, lasso, neural network, and random forest models to predict cholesterol, calculated the theoretical transition intervals for least squares and ridge regression, and evaluated the performance of the approaches. Since we expect the transition point theory to yield different conclusions when the training studies are homogeneous versus heterogeneous, we sought to empirically demonstrate this by considering two scenarios representing homogeneous and heterogeneous training studies: 1) training on $K=4$ different subsets of the same study and testing on a held out subset, and 2) training on $K=2$ different studies and testing on an independent study. In the first scenario, we randomly split the samples from \cite{qin2012metagenome} into five datasets of approximately equal size, using $K=4$ for training and the remaining one for testing. In second scenario, we used the $K=2$ datasets from \cite{qin2012metagenome} and \cite{karlsson2013gut} for training and dataset from \cite{heintz2017integrated} for testing. We used age and gene marker abundance for a selected set of gene markers as the predictors. Due to the limited sample sizes, we restricted to a small number of gene markers. In the first (second) scenario, we selected the five (twenty) gene markers most highly correlated with cholesterol in the training set (Appendix C, Supplementary Materials). The linear models were specified as $Y_{jk} = \vec{\beta^T} \vec{X_{jk}} + \epsilon_{jk}$ where $\vec{X_{jk}}$ contained 7 elements in the first scenario - intercept, age, and 5 gene markers - and 22 elements in the second scenario - intercept, age, and 20 gene markers. To estimate the transition intervals, we estimated $\mat G$ by fitting a linear mixed effects model using residual maximum likelihood (the preferred method for small sample sizes), allowing each predictor to have a random effect: $Y_{jk} = \vec{\beta^T} \vec{X_{jk}} + \vec{\gamma_k}^T \vec{X_{jk}} + \epsilon_{jk}.$
We estimated the optimal weights for least squares and ridge regression using Proposition \ref{prop1} and estimates of $\mat G$, $\sigma_\epsilon^2$, and $\vec \beta$ from the mixed effects model. We calculated the theoretical transition bounds from Theorem \ref{thm2} and compared them to the estimate of $\overline{\sigma^2}$. We used the optimal weights for ridge regression to ensemble the study-specific lasso, neural network, and random forest models. We evaluated the performance of merging and ensembling by calculating the prediction error on the test set. 

\begin{figure}[ht]
\centering
\includegraphics[width=0.8\columnwidth]{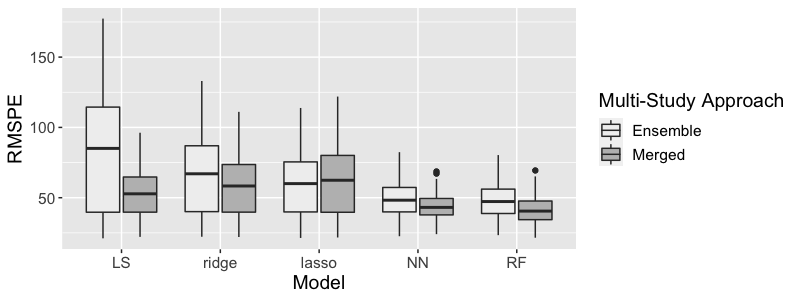}
\caption{Root mean square prediction error (RMSPE) for the first data illustration scenario with bootstrap confidence intervals. LS: least squares. NN: neural network. RF: random forest.}
\label{figure:scenario1}
\end{figure}

\begin{figure}[ht]
\centering
\includegraphics[width=0.8\columnwidth]{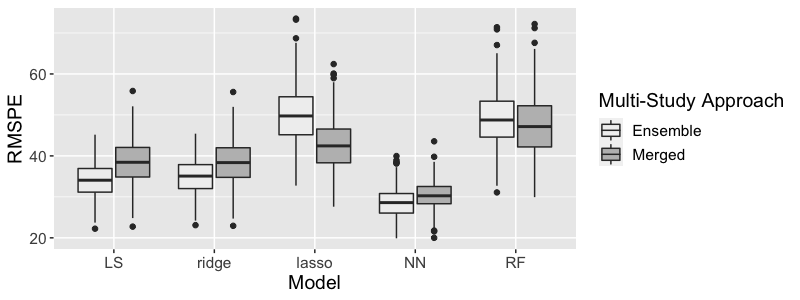}
\caption{Root mean square prediction error (RMSPE) for the second data illustration scenario with bootstrap confidence intervals.  LS: least squares. NN: neural network. RF: random forest.}
\label{figure:scenario2}
\end{figure}

In the first scenario, $\overline{\sigma^2}$ was estimated to be $0.005^2$, $\tau_{LS,1}$ was $0.266^2$, and $\tau_{R,1}$ was $0.050^2$, so merging was expected outperform ensembling. Accordingly, in the test set, the merged versions of all of the models had lower prediction error than the corresponding ensembles (Figure \ref{figure:scenario1}).
In the second scenario, $\overline{\sigma^2}$ was estimated to be $18.322^2$, $\tau_{LS,2}$ was $15.247^2$, and $\tau_{R, 2}$ was $4.423^2$, so ensembling was expected to outperform merging. Accordingly, in the test set, the ensemble versions of least squares, ridge regression, and neural network had lower prediction error than the corresponding merged learners. However, merging outperformed ensembling for lasso and merging performed similarly to ensembling for random forest (Figure \ref{figure:scenario2}) to which our theoretical results do not apply. While it is hard to generalize from a single realization, we observed that lasso and random forest both had considerably higher prediction error than the other models. Lasso and random forest both perform internal feature selection, which adds a layer of complexity that may influence their transition points. 

\section{Discussion}

The availability of large though potentially heterogeneous collections of data for training classifiers is challenging traditional approaches for training and validating prediction and classification algorithms. At the same time, it is creating opportunities for new and more general paradigms. One of these is multi-study ensemble learning, motivated by variation in the relation between predictors and outcomes across collections of similar studies. A natural benchmark for these methods is to combine all training studies to exploit the power of larger training sample sizes. In previous work \citep{patil2018training}, merged learners were shown to perform better than ensemble learners in simulations under low-heterogeneity settings. As heterogeneity increased, however, there was a transition point in the heterogeneity scale beyond which acknowledging cross-study heterogeneity became preferable, and the ensemble learners outperformed the merged learners. 

In this paper, we provided the first theoretical investigation of multi-study ensembling. Using a flexible data generating framework based on a nonparametric mixed effects model, we proved that when prediction is done using linear models, multi-study ensembling can, under certain conditions, achieve better optimality properties than merging all studies prior to training, a question that was still open. The class of linear models we have considered includes regularized and basis-expanded models, which are able to provide reasonable approximations of potentially nonlinear data generating models. We characterized the relative performance of merging and ensembling as a function of inter-study heterogeneity, deriving a closed-form expression for the transition point beyond which ensembling outperforms merging, for both least squares and ridge regression. These results are applicable to both correctly specified and misspecified linear prediction models. We confirmed the analytic results in simulations and demonstrated that when the data are generated by a mixed effects model, the least squares and ridge regression solutions can potentially serve as proxies for the transition point under other learning strategies (e.g. neural networks) for which closed-form derivation is difficult. Finally, we estimated transition bounds in cases of low and high cross-study heterogeneity in microbiome data and showed how they can be used in practice as a guide for deciding when to merge studies together in learning a prediction rule. In particular, $\overline{\sigma^2}$ can be estimated from the training data and compared to the theoretical transition points or bounds for least squares and/or ridge regression. Various methods can be used to estimate $\overline{\sigma^2}$, including maximum likelihood and method of moments-based approaches used in meta-analysis (for example, see \cite{jackson2016extending}), with the caveat that estimates may be imprecise when the number of studies is small (10 or fewer studies).

We focused on deriving analytic results for ridge regression and its variants, including least squares and penalized spline regression, because of the opportunity to pursue closed-form solutions. Other widely used methods such as lasso, neural networks, and random forests are not as easily amenable to a closed-form solution, so we used simulations to investigate whether the analytic results we derived could potentially serve as an approximation for these methods. In simulations under a linear generating model, the empirical transition points of least squares, ridge regression, lasso, and neural networks fell within a similar range, but the random forests showed very different behavior. In one setting, the ensemble version of the random forest always outperformed the merged version and there was no transition point, while in other settings, the transition for the random forest occurred much earlier than for the other models. A key difference between random forests and the other methods considered is that a single random forest is already an ensemble of decision trees, which potentially complicates the comparison of merging to ensembling. Further exploration of this issue is provided by \cite{Ramchandran:2019ga, Ramchandran2021a, ramchandran2021ensembling}. While the simulation results suggest that there are scenarios where it may be reasonable to use the transition points for ridge regression or least squares as a proxy when considering machine learning models such as lasso or neural networks, the data application shows that there are also scenarios where the transition point of lasso is dissimilar to that of ridge regression or least squares. We posit that these methods appeared similar in the simulation because feature selection was not necessary, as there were no extraneous features by design. In practice, the transition points for lasso and random forest may be affected by internal variable selection that takes place in these models. Therefore, the theoretical results presented here should not necessarily be used as an approximation of the transition points for these models.

Under Model~(\ref{model1}), fitting a correctly specified mixed effects model will generally be more efficient than both the merged and ensemble versions of least squares. However, more flexible machine learning algorithms can potentially yield better prediction accuracy than the true model. For example, in the main simulation analysis, the mixed effects model was outperformed by a neural network. 

A limitation of our derivations is that they treat the following quantities as known: the subset of predictors with random effects, the ensemble weights, and the regularization parameters for ridge regression. In practice, these are usually selected using statistical procedures that introduce additional variability. Also, we obtained closed-form transition point expressions for cases where the ensemble weighting scheme does not depend on the variances of the random effects. Such weighting schemes are generally sub-optimal, as the optimal weights given by Proposition~\ref{prop1} depend on $\mat G$. Therefore, the closed-form results are based on a conservative estimate of the maximal performance of multi-study ensembling. Another limitation is the assumption that the random effects are uncorrelated, which is often not true in practice. Furthermore, the scope of this paper is restricted to continuous outcomes. Previous studies have compared the performance of merging and ensembling for binary \citep{Ramchandran:2019ga, Ramchandran2021a} and time-to-event outcomes \citep{patil2018training} in simulations and real data applications, showing the benefits of ensembling when there is heterogeneity across studies. Developing transition point theory for more general outcomes is an interesting but challenging problem. Using asymptotic expansions to extend the theory for continuous outcomes is one potential direction for future research.

In summary, although this work is predicated upon the assumptions that cross-study heterogeneity manifests as random effects and that weights and regularization parameters are known, we believe it provides the first theoretical rationale for multi-study ensembling and a strong foundation for developing practical rules and guidelines to implement it. 

\vspace{0.5cm}
\setstretch{1.2}
\section{Supplementary Material}
The analysis code is available at {\tt https://github.com/zoeguan/transition\_point}. Supplementary material available online includes appendices with proofs, additional plots, and a notation table.

\vspace{0.5cm}
\section{Competing interests}
No competing interest is declared.

\vspace{0.5cm}
\section{Author contributions statement}

Z.G., G.P., and P.P. conceived the experiments, Z.G. conducted the experiments, and Z.G., G.P., and P.P. wrote and reviewed the manuscript. 

\vspace{0.5cm}
\section{Acknowledgments}
The authors thank Lorenzo Trippa and Boyu Ren for helpful discussions. Work supported by NSF-DMS grants (1810829, 2113707) and NSERC PGSD3 - 502362 - 2017.

\vspace{0.5cm}
\bibliographystyle{agsm}
\bibliography{reference}

\newpage
\setcounter{page}{1}

\begin{appendices}

\section*{Appendix A: Calculations and Proofs}

\subsection{Covariance Matrices and MSPEs}

The covariances and expectations are all conditional on the predictors. 

\begin{flalign*}
Cov(\vec{Y_k}) &= Cov(\mat f(\mat{X_{k}}) + \mat{Z_{k}}\vec{\gamma_k} + \vec{\epsilon_{k}}) &\\
&= \mat{Z_k} Cov(\vec{\gamma_k}) \mat{Z_k}^T + Cov(\vec{\epsilon_k}) &\\
&= \mat{Z_k} \mat{G} \mat{Z_k}^T + \sigma_\epsilon^2 \mat{I_{N_k}}
\end{flalign*}

\begin{flalign*}
Cov(\mat{\tilde{X}_0} \vec{\hat{\beta}_{k}} ) &= \mat{\tilde{X}_0} Cov(\vec{\hat{\beta}_{k}}) \mat{\tilde{X}_0}^T &\\
&= \mat{\tilde{X}_0} \mat{\tilde{R}_k}^{-1} \mat{\tilde{X}_k}^T Cov(\vec{Y_k}) \mat{\tilde{X}_k} \mat{\tilde{R}_k}^{-1} \mat{\tilde{X}_0}^T \\
&= \mat{\tilde{X}_0} \mat{\tilde{R}_k}^{-1} \mat{\tilde{X}_k}^T \mat{Z_k} \mat{G} \mat{Z_k}^T \mat{\tilde{X}_k} \mat{\tilde{R}_k}^{-1} \mat{\tilde{X}_0}^T + \sigma_\epsilon^2 \mat{\tilde{X}_0} \mat{\tilde{R}_k}^{-1} \mat{\tilde{X}_k}^T \mat{\tilde{X}_k} \mat{\tilde{R}_k}^{-1} \mat{\tilde{X}_0}^T
\end{flalign*}

\begin{flalign*}
Cov(\mat{\tilde{X}_0} \vec{\hat{\beta}_{ens}} ) &= \sum_{k=1}^K w_k^2 Cov(\mat{\tilde{X}_0} \vec{\hat{\beta}_{k}} ) &\\
&= \sum_{k=1}^K w_k^2 \mat{\tilde{X}_0} \mat{\tilde{R}_k}^{-1} \mat{\tilde{X}_k}^T \mat{Z_k} \mat{G} \mat{Z_k}^T \mat{\tilde{X}_k} \mat{\tilde{R}_k}^{-1} \mat{\tilde{X}_0}^T + \sigma_\epsilon^2 \sum_{k=1}^K w_k^2 \mat{\tilde{X}_0} \mat{\tilde{R}_k}^{-1} \mat{\tilde{X}_k}^T \mat{\tilde{X}_k} \mat{\tilde{R}_k}^{-1} \mat{\tilde{X}_0}^T 
\end{flalign*}

\begin{flalign*}
\vec{b_{ens}} = Bias(\mat{\tilde{X}_0} \vec{\hat{\beta}_{ens}} ) &= \sum_{k=1}^K w_k E[\mat{\tilde{X}_0} \vec{\hat{\beta}_{k}}] - \vec f(\mat{X_0}) &\\
&= \sum_{k=1}^K w_k \mat{\tilde{X}_0} \mat{\tilde{R}_k}^{-1} \mat{\tilde{X}_k}^T  \vec f(\mat{X_k}) - \vec f(\mat{X_0})
\end{flalign*}

\begin{flalign*}
Cov(\mat{\tilde{X}_0} \vec{\hat{\beta}_{merge}} ) &= \mat{\tilde{X}_0} Cov(\vec{\hat{\beta}_{merge}}) \mat{\tilde{X}_0}^T &\\
&= \mat{\tilde{X}_0}   Cov(\mat{\tilde{R}}^{-1} \sum_{k=1}^K \mat{\tilde{X}_k}^T \vec Y_k) \mat{\tilde{X}_0}^T &\\
&= \mat{\tilde{X}_0} \mat{\tilde{R}}^{-1} \sum_{k=1}^K \mat{\tilde{X}_k}^T Cov(\vec Y_k) \mat{\tilde{X}_k} \mat{\tilde{R}}^{-1} \mat{\tilde{X}_0}^T &\\
&= \mat{\tilde{X}_0} \mat{\tilde{R}}^{-1} \sum_{k=1}^K \mat{\tilde{X}_k}^T \mat{Z_k} \mat{G} \mat{Z_k}^T  \mat{\tilde{X}_k} \mat{\tilde{R}}^{-1} \mat{\tilde{X}_0}^T + \sigma_\epsilon^2 \mat{\tilde{X}_0} \mat{\tilde{R}}^{-1} \sum_{k=1}^K \mat{\tilde{X}_k}^T\mat{\tilde{X}_k} \mat{\tilde{R}}^{-1} \mat{\tilde{X}_0}^T   &\\
&= \mat{\tilde{X}_0} \mat{\tilde{R}}^{-1} \sum_{k=1}^K \mat{\tilde{X}_k}^T \mat{Z_k} \mat{G} \mat{Z_k}^T  \mat{\tilde{X}_k} \mat{\tilde{R}}^{-1} \mat{\tilde{X}_0}^T + \sigma_\epsilon^2 \mat{\tilde{X}_0} \mat{\tilde{R}}^{-1} \mat{\tilde{X}}^T\mat{\tilde{X}} \mat{\tilde{R}}^{-1} \mat{\tilde{X}_0}^T
\end{flalign*}

\begin{flalign*}
\vec{b_{merge}}=Bias(\mat{\tilde{X}_0} \vec{\hat{\beta}_{merge}} ) &= E[\mat{\tilde{X}_0} \vec{\hat{\beta}_{merge}}] - \vec f(\mat{X_0}) &\\
&= \mat{\tilde{X}_0} \mat{\tilde{R}}^{-1} \mat{\tilde{X}}^T \vec f(\mat{X}) - \vec f(\mat{X_0})
\end{flalign*}

\begin{flalign*}
E[\| \mat {Y_0} - \mat{\tilde{X}_0} \vec{\hat\beta_{ens}} \|_2^2 | \mat{{X}_0}] &= tr(Cov(\mat{\tilde{X}_0} \vec{\hat{\beta}_{ens}} )) +  \vec{b_{ens}}^T \vec{b_{ens}} + E[(\vec {Y_0} - \vec f(\mat{X_0}))^2] &\\
&= \sum_{k=1}^K w_k^2  tr(\mat{\tilde{X}_0} \mat{\tilde{R}_k}^{-1} \mat{\tilde{X}_k}^T \mat{Z_k}  \mat{G} \mat{Z_k}^T \mat{\tilde{X}_k} \mat{\tilde{R}_k}^{-1} \mat{\tilde{X}_0}^T) + \\
& \phantom{==} \sigma_\epsilon^2 \sum_{k=1}^K  tr( \mat{\tilde{X}_0} \mat{\tilde{R}_k}^{-1} \mat{\tilde{X}_k}^T \mat{\tilde{X}_k} \mat{\tilde{R}_k}^{-1} \mat{\tilde{X}_0}^T w_k^2) + \vec{b_{ens}}^T \vec{b_{ens}} + E[(\vec {Y_0} - \vec f(\mat{X_0}))^2] &\\
&= \sum_{k=1}^K w_k^2  tr(\mat{G} \mat{Z_k}^T \mat{\tilde{X}_k} \mat{\tilde{R}_k}^{-1} \mat{\tilde{X}_0}^T\mat{\tilde{X}_0} \mat{\tilde{R}_k}^{-1} \mat{\tilde{X}_k}^T \mat{Z_k} ) + \\
& \phantom{==} \sigma_\epsilon^2 \sum_{k=1}^K w_k^2 tr( \mat{\tilde{X}_0} \mat{\tilde{R}_k}^{-1} \mat{\tilde{X}_k}^T \mat{\tilde{X}_k} \mat{\tilde{R}_k}^{-1} \mat{\tilde{X}_0}^T) + \vec{b_{ens}}^T \vec{b_{ens}} + E[(\vec {Y_0} - \vec f(\mat{X_0}))^2] &\\
&= \sum_{i=1}^Q  \sigma_i^2  \sum_{k=1}^K w_k^2 (\mat{Z_k}^T \mat{\tilde{X}_k} \mat{\tilde{R}_k}^{-1} \mat{\tilde{X}_0}^T\mat{\tilde{X}_0} \mat{\tilde{R}_k}^{-1} \mat{\tilde{X}_k}^T \mat{Z_k} )_{ii} + \\
& \phantom{==} \sigma_\epsilon^2 \sum_{k=1}^K w_k^2 tr( \mat{\tilde{X}_0} \mat{\tilde{R}_k}^{-1} \mat{\tilde{X}_k}^T \mat{\tilde{X}_k} \mat{\tilde{R}_k}^{-1} \mat{\tilde{X}_0}^T) + \vec{b_{ens}}^T \vec{b_{ens}} + E[(\vec {Y_0} - \vec f(\mat{X_0}))^2] &\\
&= \sum_{d=1}^D \sigma_{(d)}^2 \sum_{i:\sigma_i^2=\sigma_{(d)}^2} \sum_{k=1}^K w_k^2 (\mat{Z_k}^T \mat{\tilde{X}_k} \mat{\tilde{R}_k}^{-1} \mat{\tilde{X}_0}^T\mat{\tilde{X}_0} \mat{\tilde{R}_k}^{-1} \mat{\tilde{X}_k}^T \mat{Z_k} )_{ii} + \\
& \phantom{==} \sigma_\epsilon^2 \sum_{k=1}^K w_k^2 tr( \mat{\tilde{X}_0} \mat{\tilde{R}_k}^{-1} \mat{\tilde{X}_k}^T \mat{\tilde{X}_k} \mat{\tilde{R}_k}^{-1} \mat{\tilde{X}_0}^T) + \vec{b_{ens}}^T \vec{b_{ens}} + E[(\vec {Y_0} - \vec f(\mat{X_0}))^2] &\\
&= \sum_{d=1}^D \sigma_{(d)}^2 \sum_{i:\sigma_i^2=\sigma_{(d)}^2} \sum_{k=1}^K w_k^2 (\mat{Z_k}^T \mat{\tilde{X}_k} \mat{\tilde{R}_k}^{-1} \mat{\tilde{X}_0}^T\mat{\tilde{X}_0} \mat{\tilde{R}_k}^{-1} \mat{\tilde{X}_k}^T \mat{Z_k} )_{ii} + \\
& \phantom{==} \sigma_\epsilon^2 \sum_{k=1}^K w_k^2 tr( \mat{\tilde{X}_k} \mat{\tilde{R}_k}^{-1} \mat{\tilde{X}_0}^T \mat{\tilde{X}_0} \mat{\tilde{R}_k}^{-1} \mat{\tilde{X}_k}^T ) + \vec{b_{ens}}^T \vec{b_{ens}} + E[(\vec {Y_0} - \vec f(\mat{X_0}))^2] &\\
&= \sum_{d=1}^D \sigma_{(d)}^2 \sum_{i:\sigma_i^2=\sigma_{(d)}^2} \sum_{k=1}^K w_k^2 (\mat{Z_k}^T \mat{\tilde{X}_k} \mat{\tilde{R}_k}^{-1} \mat{\tilde{X}_0}^T\mat{\tilde{X}_0} \mat{\tilde{R}_k}^{-1} \mat{\tilde{X}_k}^T \mat{Z_k} )_{ii} + \\
& \phantom{==} \sigma_\epsilon^2 \sum_{k=1}^K w_k^2 tr( \mat{\tilde{A}_k} ) + \vec{b_{ens}}^T \vec{b_{ens}} + E[(\vec {Y_0} - \vec f(\mat{X_0}))^2] &\\
\end{flalign*}

\begin{flalign*}
E[\| \vec {Y_0} - \mat{\tilde{X}_0} \vec{\hat\beta_{merge}} \|_2^2 | \mat{{X}_0}] &= tr(Cov(\mat{\tilde{X}_0} \vec{\hat{\beta}_{merge}} )) +  \vec{b_{merge}}^T \vec{b_{merge}} + E[(\vec {Y_0} - \vec f(\mat{X_0}))^2] &\\
&= tr(\mat{\tilde{X}_0} \mat{\tilde{R}}^{-1} \sum_{k=1}^K \mat{\tilde{X}_k}^T \mat{Z_k} \mat{G} \mat{Z_k}^T  \mat{\tilde{X}_k} \mat{\tilde{R}}^{-1} \mat{\tilde{X}_0}^T) + \\
& \phantom{==} \sigma_\epsilon^2 tr(\mat{\tilde{X}_0} \mat{\tilde{R}}^{-1} \mat{\tilde{X}}^T\mat{\tilde{X}} \mat{\tilde{R}}^{-1} \mat{\tilde{X}_0}^T) + \vec{b_{merge}}^T \vec{b_{merge}} + E[(\vec {Y_0} - \vec f(\mat{X_0}))^2] &\\
&=  tr(\mat{G} \sum_{k=1}^K \mat{Z_k}^T  \mat{\tilde{X}_k} \mat{\tilde{R}}^{-1} \mat{\tilde{X}_0}^T \mat{\tilde{X}_0} \mat{\tilde{R}}^{-1} \mat{\tilde{X}_k}^T \mat{Z_k} ) + \\
& \phantom{==} \sigma_\epsilon^2 tr(\mat{\tilde{X}_0} \mat{\tilde{R}}^{-1} \mat{\tilde{X}}^T\mat{\tilde{X}} \mat{\tilde{R}}^{-1} \mat{\tilde{X}_0}^T) + \vec{b_{merge}}^T \vec{b_{merge}} + E[(\vec {Y_0} - \vec f(\mat{X_0}))^2] &\\
&= \sum_{i=1}^Q \sigma_i^2 \sum_{k=1}^K (\mat{Z_k}^T  \mat{\tilde{X}_k} \mat{\tilde{R}}^{-1} \mat{\tilde{X}_0}^T \mat{\tilde{X}_0} \mat{\tilde{R}}^{-1} \mat{\tilde{X}_k}^T \mat{Z_k} )_{ii} + \\
& \phantom{==} \sigma_\epsilon^2 tr(\mat{\tilde{X}_0} \mat{\tilde{R}}^{-1} \mat{\tilde{X}}^T\mat{\tilde{X}} \mat{\tilde{R}}^{-1} \mat{\tilde{X}_0}^T) + \vec{b_{merge}}^T \vec{b_{merge}} + E[(\vec {Y_0} - \vec f(\mat{X_0}))^2] &\\
&= \sum_{d=1}^D \sigma_{(d)}^2 \sum_{i:\sigma_i^2=\sigma_{(d)}^2} \sum_{k=1}^K (\mat{Z_k}^T  \mat{\tilde{X}_k} \mat{\tilde{R}}^{-1} \mat{\tilde{X}_0}^T \mat{\tilde{X}_0} \mat{\tilde{R}}^{-1} \mat{\tilde{X}_k}^T \mat{Z_k} )_{ii} + \\
& \phantom{==} \sigma_\epsilon^2 tr(\mat{\tilde{X}}  \mat{\tilde{R}}^{-1}\mat{\tilde{X}_0}^T\mat{\tilde{X}_0} \mat{\tilde{R}}^{-1} \mat{\tilde{X}}^T) + \vec{b_{merge}}^T \vec{b_{merge}} + E[(\vec {Y_0} - \vec f(\mat{X_0}))^2] &\\
&= \sum_{d=1}^D \sigma_{(d)}^2 \sum_{i:\sigma_i^2=\sigma_{(d)}^2} \sum_{k=1}^K (\mat{Z_k}^T  \mat{\tilde{X}_k} \mat{\tilde{R}}^{-1} \mat{\tilde{X}_0}^T \mat{\tilde{X}_0} \mat{\tilde{R}}^{-1} \mat{\tilde{X}_k}^T \mat{Z_k} )_{ii} + \\
& \phantom{==} \sigma_\epsilon^2 tr(\mat{\tilde{X}}^T \mat{\tilde{X}}  \mat{\tilde{R}}^{-1}\mat{\tilde{X}_0}^T\mat{\tilde{X}_0} \mat{\tilde{R}}^{-1}) + \vec{b_{merge}}^T \vec{b_{merge}} + E[(\vec {Y_0} - \vec f(\mat{X_0}))^2]&\\
&= \sum_{d=1}^D \sigma_{(d)}^2 \sum_{i:\sigma_i^2=\sigma_{(d)}^2} \sum_{k=1}^K (\mat{Z_k}^T  \mat{\tilde{X}_k} \mat{\tilde{R}}^{-1} \mat{\tilde{X}_0}^T \mat{\tilde{X}_0} \mat{\tilde{R}}^{-1} \mat{\tilde{X}_k}^T \mat{Z_k} )_{ii} + \\
& \phantom{==} \sigma_\epsilon^2 tr(\sum_k \mat{\tilde{X_k}}^T \mat{\tilde{X_k}}  \mat{\tilde{R}}^{-1}\mat{\tilde{X}_0}^T\mat{\tilde{X}_0} \mat{\tilde{R}}^{-1}) + \vec{b_{merge}}^T \vec{b_{merge}} + E[(\vec {Y_0} - \vec f(\mat{X_0}))^2]&\\
&= \sum_{d=1}^D \sigma_{(d)}^2 \sum_{i:\sigma_i^2=\sigma_{(d)}^2} \sum_{k=1}^K (\mat{Z_k}^T  \mat{\tilde{X}_k} \mat{\tilde{R}}^{-1} \mat{\tilde{X}_0}^T \mat{\tilde{X}_0} \mat{\tilde{R}}^{-1} \mat{\tilde{X}_k}^T \mat{Z_k} )_{ii} + \\
& \phantom{==} \sigma_\epsilon^2 \sum_k tr( \mat{\tilde{X_k}}  \mat{\tilde{R}}^{-1}\mat{\tilde{X}_0}^T\mat{\tilde{X}_0} \mat{\tilde{R}}^{-1} \mat{\tilde{X_k}}^T ) + \vec{b_{merge}}^T \vec{b_{merge}} + E[(\vec {Y_0} - \vec f(\mat{X_0}))^2]&\\
&= \sum_{d=1}^D \sigma_{(d)}^2 \sum_{i:\sigma_i^2=\sigma_{(d)}^2} \sum_{k=1}^K (\mat{Z_k}^T  \mat{\tilde{X}_k} \mat{\tilde{R}}^{-1} \mat{\tilde{X}_0}^T \mat{\tilde{X}_0} \mat{\tilde{R}}^{-1} \mat{\tilde{X}_k}^T \mat{Z_k} )_{ii} + \\
& \phantom{==} \sigma_\epsilon^2 \sum_k tr( \mat{\tilde{C}_k}) + \vec{b_{merge}}^T \vec{b_{merge}} + E[(\vec {Y_0} - \vec f(\mat{X_0}))^2]&\\
\end{flalign*}

\subsection{Proof of Theorem 2}

Theorems 1 is a special case of Theorem 2.

Let $a_d = \sum\limits_{i:\sigma_i^2=\sigma_{(d)}^2} \sum_{k=1}^K \left( (\mat{Z_k}^T  \mat{\tilde{X}_k} \mat{\tilde{R}}^{-1} \mat{\tilde{X}_0}^T \mat{\tilde{X}_0} \mat{\tilde{R}}^{-1}  \mat{\tilde{X}_k}^T \mat{Z_k} )_{ii} - w_k^2 (\mat{Z_k}^T \mat{\tilde{X}_k} \mat{\tilde{R}_k}^{-1} \mat{\tilde{X}_0}^T\mat{\tilde{X}_0} \mat{\tilde{R}_k}^{-1} \mat{\tilde{X}_k}^T \mat{Z_k} )_{ii} \right)$.

Let $c = \sigma_\epsilon^2 \left(\sum\limits_{k=1}^K w_k^2 tr( \mat{\tilde{X}_0} \mat{\tilde{R}_k}^{-1} \mat{\tilde{X}_k}^T \mat{\tilde{X}_k} \mat{\tilde{R}_k}^{-1} \mat{\tilde{X}_0}^T) - tr(\mat{\tilde{X}_0} \mat{\tilde{R}}^{-1} \mat{\tilde{X}}^T\mat{\tilde{X}} \mat{\tilde{R}}^{-1} \mat{\tilde{X}_0}^T) \right) + \vec{b_{ens}}^T \vec{b_{ens}} - \vec{b_{merge}}^T \vec{b_{merge}}$. \\ 

Since
\begin{flalign}
& E[\| \vec {Y_0} - \mat{\tilde{X}_0} \vec{\hat\beta_{merge}} \|_2^2 | \mat{{X}_0}] \geq
E[\| \mat {Y_0} - \mat{\tilde{X}_0} \vec{\hat\beta_{ens}} \|_2^2 | \mat{{X}_0}] \iff \sum_{d=1}^D \sigma_{(d)}^2 a_d        \geq         c &
\end{flalign}
and
\begin{flalign}
\left(\min_d \frac{a_d}{J_d}\right) \sum_{d=1}^D \sigma_{(d)}^2 J_d    \leq   \sum_{d=1}^D \sigma_{(d)}^2 a_d   \leq   \left(\max_d \frac{a_d}{J_d} \right) \sum_{d=1}^D \sigma_{(d)}^2 J_d \quad, 
\end{flalign}
it follows that
{\begin{flalign*}
& \overline{\sigma^2} = \frac{\sum\limits_{d=1}^D \sigma_{(d)}^2 J_d} {P}      \leq      \frac{c}   {P \max\limits_d \frac{a_d}{J_d}} = \tau_1   &\\
& \implies \quad   \sum_{d=1}^D \sigma_{(d)}^2 a_d \leq \max\limits_d \frac{a_d}{J_d} \sum\limits_{d=1}^D \sigma_{(d)}^2 J_d \leq c    &\\
& \iff \quad   E[\| \vec {Y_0} - \mat{\tilde{X}_0} \vec{\hat\beta_{merge}} \|_2^2 | \mat{{X}_0}] \leq
E[\| \mat {Y_0} - \mat{\tilde{X}_0} \vec{\hat\beta_{ens}} \|_2^2 | \mat{{X}_0}]. &
\end{flalign*}}

Similarly,
{\begin{flalign*}
& \overline{\sigma^2} = \frac{\sum\limits_{d=1}^D \sigma_{(d)}^2 J_d} {P}      \geq      \frac{c}   {P \max\limits_d \frac{a_d}{J_d}} = \tau_2   &\\
& \implies \quad   \sum_{d=1}^D \sigma_{(d)}^2 a_d \geq \min\limits_d \frac{a_d}{J_d} \sum\limits_{d=1}^D \sigma_{(d)}^2 J_d \geq c   &\\
& \iff \quad    E[\| \vec {Y_0} - \mat{\tilde{X}_0} \vec{\hat\beta_{merge}} \|_2^2 | \mat{{X}_0}] \geq
E[\| \mat {Y_0} - \mat{\tilde{X}_0} \vec{\hat\beta_{ens}} \|_2^2 | \mat{{X}_0}]. &
\end{flalign*}}

\subsection{Optimal Weights}

\textbf{Ridge Regression.} For ridge regression, let 

\[v_k = tr(Cov(\mat{\tilde{X}_0}  \vec{\hat{\beta}_{k}} )) = tr(\mat{\tilde{X}_0} \mat{\tilde{R}_k}^{-1} \mat{\tilde{X}_k}^T \mat{Z_k} \mat{G} \mat{Z_k}^T \mat{\tilde{X}_k} \mat{\tilde{R}_k}^{-1} \mat{\tilde{X}_0}^T + \sigma_\epsilon^2 \mat{\tilde{X}_0} \mat{\tilde{R}_k}^{-1} \mat{\tilde{X}_k}^T \mat{\tilde{X}_k} \mat{\tilde{R}_k}^{-1} \mat{\tilde{X}_0}^T )\]
\[\mat{b_k} =  \mat{\tilde{X}_0} \mat{\tilde{R}_k}^{-1} \mat{\tilde{X}_k}^T  \vec f(\mat{X_k}) - \vec f(\mat{X_0}).\]
We want to minimize
\[ \sum_k w_k^2 v_k + \left(\sum_k w_k \mat{b_k}\right)^T \left(\sum_k w_k \mat{b_k}\right)  \]
subject to the constraint $\sum_k w_k = 1$.

Using Lagrange multipliers, we get the system of equations
\begin{flalign*}
2w_k (v_k+ \mat{b_k}^T \mat{b_k}) + 2\sum_{j \neq k} w_j \mat{b_j}^T \mat{b_k} &= \alpha  \tag{for $k=1,\dots,K$}\\
\sum_k w_k &= 1.
\end{flalign*}

The solution to the system is
\begin{flalign}
\begin{bmatrix} \vec{w} \\ \alpha \end{bmatrix}= \begin{bmatrix} \mat E & -\vec j \\ \vec j^T & 0  \end{bmatrix}^{-1} \begin{bmatrix} \vec 0 \\ 1 \end{bmatrix}
\end{flalign}
where the entries of $\mat E$ are $(\mat E)_{kk} = (v_k+ \mat{b_k}^T \mat{b_k}) $ and $(\mat E)_{jk} =  \mat{b_j}^T \mat{b_k}$ for $j \neq k$ and $\vec j = (1, \dots, 1)^T \in \mathds{R}^K$.

\begin{flalign}
\vec{w} = \mat{E}^{-1} \vec J (\vec j^T \mat{E}^{-1} \vec j)^{-1}
\end{flalign}

\vspace{0.5cm}

\textbf{Least Squares Under a Linear Generating Model. } Let $\mat{R_k} = \mat{X_k}^T \mat{X_k}$. For least squares, we want to minimize 
\[ \sum_{k=1}^K w_k^2 tr(\mat{G} \mat{Z_0}^T \mat{Z_0}) + \sigma_\epsilon^2 \sum_{k=1}^K w_k^2 tr(\mat{R_{k}}^{-1} \mat{R_0}) \]
subject to the constraint $\sum_k w_k = 1$.

Using Lagrange multipliers, we get the system of equations
\begin{flalign*}
2 w_k \left( tr(\mat{G} \mat{Z_0}^T \mat{Z_0}) + \sigma_\epsilon^2 tr (\mat{R_{k}}^{-1} \mat{R_{0}}) \right) = \alpha \tag{for $k=1,\dots,K$; $\alpha$ is the Lagrange multiplier} \\
\sum_k w_k = 1
\end{flalign*}
which is solved by
\[ w_k = \dfrac{\left(tr(\mat{G} \mat{Z_0}^T \mat{Z_0})+ \sigma_\epsilon^2 tr (\mat{R_{k}}^{-1} \mat{R_{0}}) \right)^{-1}}{\sum_k \left( tr(\mat{G} \mat{Z_0}^T \mat{Z_0})+ \sigma_\epsilon^2 tr (\mat{R_{k}}^{-1} \mat{R_{0}}) \right)^{-1}}\]

\vspace{0.3cm}
Plugging the optimal weights into the MSPE of the ensemble learner, $tr(Cov({\mat{X_0}} \vec{{\hat\beta}_{LS,ens}}))$, we get
\[ tr(Cov({\mat{X_0}} \vec{{\hat\beta}_{LS,ens}} )) = \frac{1}{\sum_k \left( tr(\mat{G} \mat{Z_0}^T \mat{Z_0})+ \sigma_\epsilon^2 tr (\mat{R_{k}}^{-1} \mat{R_{0}}) \right)^{-1}}\]

In the equal variances setting,
\[ tr(Cov({\mat{X_0}} \vec{{\hat\beta}_{LS,ens}} )) = \frac{1}{\sum_k \left( \sigma^2 tr( \mat{Z_0}^T \mat{Z_0})+ \sigma_\epsilon^2 tr (\mat{R_{k}}^{-1} \mat{R_{0}}) \right)^{-1}}\]
so $tr(Cov({\mat{X_0}} \vec{{\hat\beta}_{LS,ens}} ))$ is approximately linear in $\sigma^2$ when $tr(\mat{R_{k}}^{-1} \mat{R_{0}})$ is similar across studies or when ${\sigma^2}$ is large.

\section*{Appendix B: Additional Plots}

\begin{figure}[H]
\centering
\includegraphics[width=0.6\columnwidth]{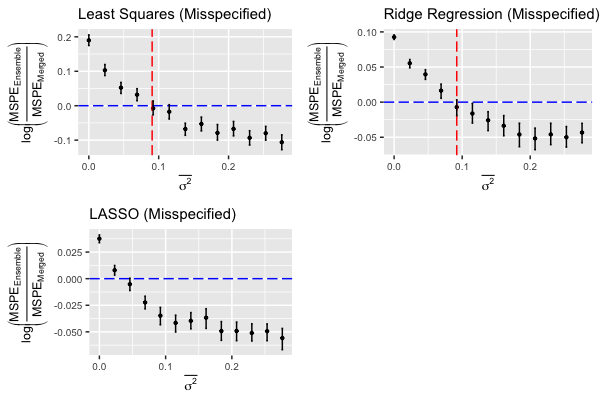}
\caption{Relative performance of multi-study ensembling and merging as a function of heterogeneity in main simulation scenario for misspecified linear models fit using the original predictors without basis expansion. MSPE: mean squared prediction error. The vertical dashed lines correspond to the transition points calculated using Theorem 1.}
\label{figure:thm1_misspec}
\end{figure}

\begin{figure}[H]
\centering
\includegraphics[width=0.6\columnwidth]{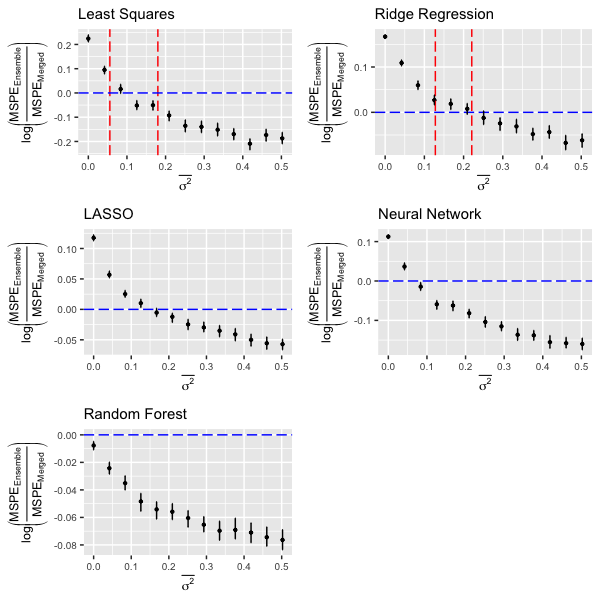}
\caption{Relative performance of multi-study ensembling and merging as a function of heterogeneity when $K=5$ and the random effects have unequal variances. MSPE: mean squared prediction error. The vertical dashed lines correspond to the bounds calculated using Theorem 2.}
\label{figure:thm2}
\end{figure}

\begin{figure}[H]
\centering
\includegraphics[width=0.6\columnwidth]{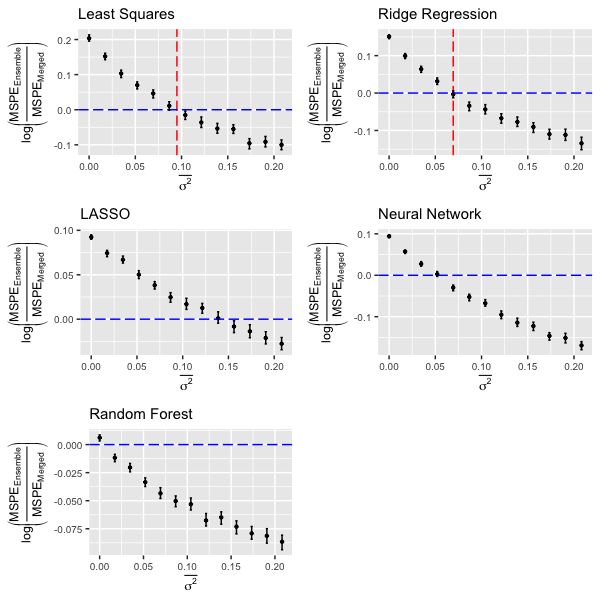}
\caption{Relative performance of multi-study ensembling and merging as a function of heterogeneity when $K=10$, $N_k=50$, and the random effects have equal variances. MSPE: mean squared prediction error. The vertical dashed lines correspond to the transition points calculated using Theorem 1.}
\label{figure:thm1_10}
\end{figure}

\begin{figure}[H]
\centering
\includegraphics[width=0.6\columnwidth]{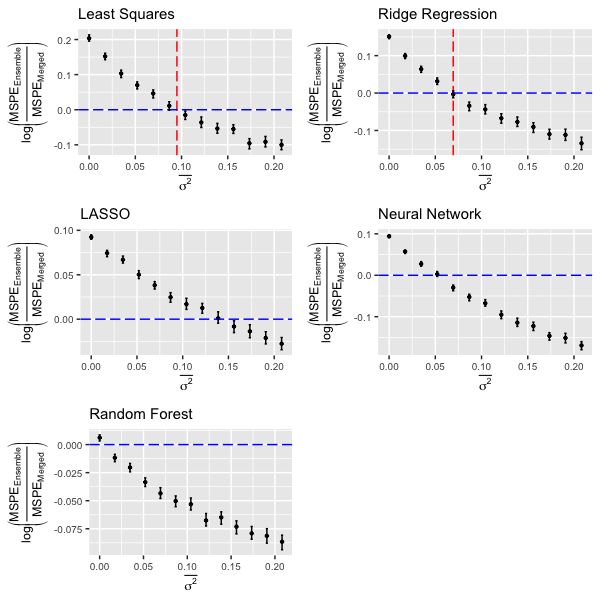}
\caption{Relative performance of multi-study ensembling and merging as a function of heterogeneity when $K=15$, $N_k=50$, and the random effects have equal variances. MSPE: mean squared prediction error. The vertical dashed lines correspond to the transition points calculated using Theorem 1.}
\label{figure:thm1_15}
\end{figure}

\begin{figure}[H]
\centering
\includegraphics[width=0.6\columnwidth]{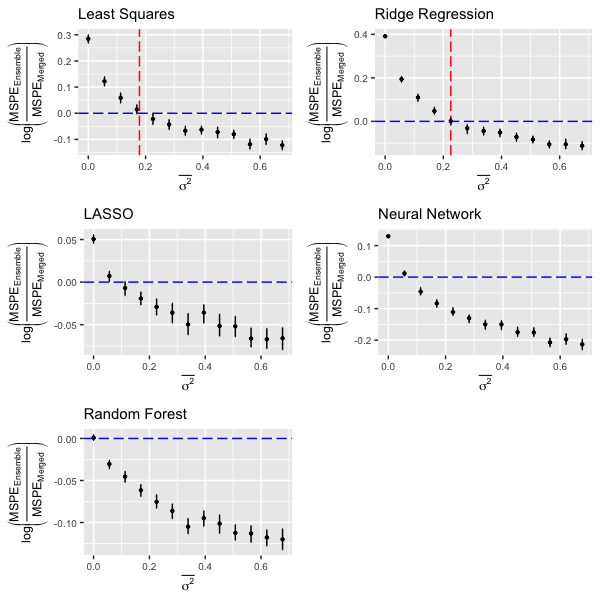}
\caption{Relative performance of multi-study ensembling and merging as a function of heterogeneity when $K=5$, $N_k=100$, and the random effects have equal variances. MSPE: mean squared prediction error. The vertical dashed lines correspond to the transition points calculated using Theorem 1.}
\label{figure:thm1_n100}
\end{figure}

\begin{figure}[H]
\centering
\includegraphics[width=0.6\columnwidth]{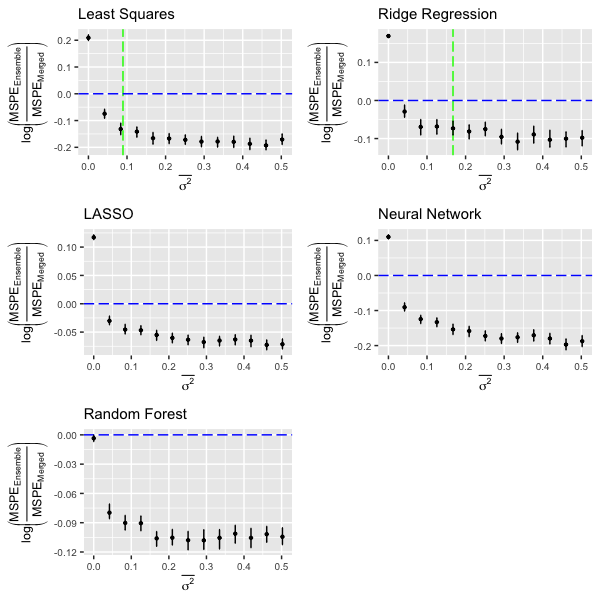}
\caption{Relative performance of multi-study ensembling and merging as a function of heterogeneity when $K=5$, $N_k=50$, and the random effects have equal variances $\sigma^2$ and are positively correlated with correlation $0.25\sigma^2$. MSPE: mean squared prediction error. The vertical dashed lines correspond to the transition points calculated using Theorem 1 under uncorrelated random effects.}
\label{figure:thm1_poscorr}
\end{figure}

\begin{figure}[H]
\centering
\includegraphics[width=0.6\columnwidth]{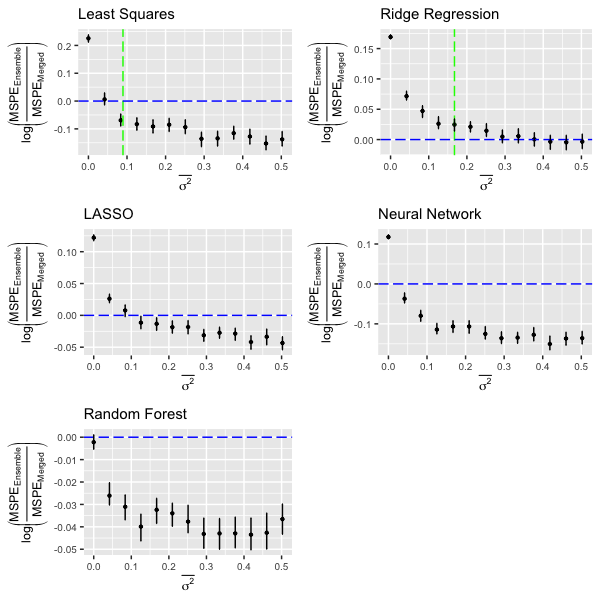}
\caption{Relative performance of multi-study ensembling and merging as a function of heterogeneity when $K=5$, $N_k=50$, and the random effects have equal variances $\sigma^2$ and are negatively correlated with correlation $-0.25\sigma^2$. MSPE: mean squared prediction error. The vertical dashed lines correspond to the transition points calculated using Theorem 1 under uncorrelated random effects.}
\label{figure:thm1_negcorr}
\end{figure}

\begin{figure}[H]
\includegraphics[width=0.6\columnwidth]{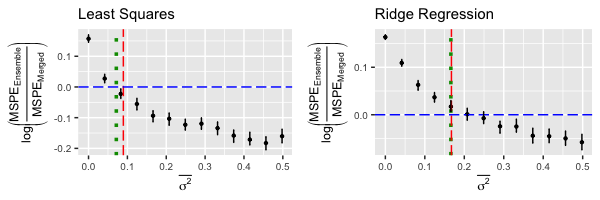}
\caption{Relative performance of multi-study ensembling and merging as a function of heterogeneity when $K=5$, $N_k=50$, $P=10$, $Q=5$, and the random effects have equal variances. MSPE: mean squared prediction error. We used the same simulated data as described in the main simulation scenario. However, instead of using equal weights in the ensembles, we estimated the optimal weights by plugging in estimates of $\mat G$, $\sigma_\epsilon^2$, and $\vec f$ from a correctly specified linear mixed effects model. The vertical dotted lines correspond to the transition points based on the true optimal weights. The vertical dashed lines correspond to the transition points based on equal weights. The empirical transition point occurs at the value of $\overline{\sigma^2}$ where the log ratio of the prediction errors for ensembling and merging is 0.}
\label{figure:thm1_opt}
\end{figure}

\begin{table}[hptb]
\centering
\begin{tabular}{ccc}
 & $\overline{\sigma^2}$ & Proportion Correctly Predicted \\ 
  \hline
1 & 0.00 & 1.00 \\ 
  2 & 0.10 & 0.93 \\ 
  3 & 0.20 & 0.56 \\ 
  4 & 0.30 & 0.49 \\ 
  5 & 0.40 & 0.52 \\ 
  6 & 0.50 & 0.58 \\ 
  7 & 0.60 & 0.68 \\ 
  8 & 0.70 & 0.67 \\ 
  9 & 0.80 & 0.68 \\ 
   \hline
\end{tabular}
\label{table:cv_hyperparams}
\caption{Comparison of the theoretical and empirical performance of merging and multi-study ensembling for ridge regression across different levels of heterogeneity when the hyperparameters are chosen using 10-fold cross-validation. The simulation setup was the same as in the main simulation scenario ($K=5$, $N_k=50$, $P=10$, $Q=5$, and the random effects have equal variances), except we varied $\overline{\sigma^2}$ from 0 to 0.8 and instead of fixing the hyperparameters $\lambda$ and $\lambda_k$ across all runs, in each run the hyperparameters were chosen via cross-validation using the training set. We performed 1000 runs per value of $\overline{\sigma^2}$. The ``Proportion Correctly Predicted" column shows the proportion of runs for which the theoretical transition point $\tau$ correctly predicted whether merging would outperform ensembling. In each run, we calculated the theoretical transition point $\tau$ by treating the cross-validated hyperparameters $\lambda$ and $\lambda_k$ as fixed and checked whether $\overline{\sigma^2} < \tau$ ($\overline{\sigma^2} < \tau$ indicates that merging should outperform ensembling and $\overline{\sigma^2} > \tau$ indicates that ensembling should outperform merging).}
\end{table}

\newpage
\section*{Appendix C: Gene Markers Used in Data Example}

The gene markers selected for the first scenario in the data example were:
\begin{itemize}
    \item gi.512436192.ref.NZ\_KE159480.1..202016.203626 
    \item gi.512436192.ref.NZ\_KE159480.1..c204954.204127 
    \item gi.512436192.ref.NZ\_KE159480.1..213329.214486 \item gi.512436192.ref.NZ\_KE159480.1..c198594.196087 
    \item gi.512436192.ref.NZ\_KE159480.1..c204130.203732
\end{itemize}

The gene markers selected for the second scenario were:
\begin{itemize}
    \item gi.479140210.ref.NC\_021010.1..1710650.1711408
\item gi.238922432.ref.NC\_012781.1..1198506.1198766
\item gi.238922432.ref.NC\_012781.1..1209760.1211007
\item gi.479213596.ref.NC\_021044.1..c2622668.2622303
\item gi.238922432.ref.NC\_012781.1..c2899618.289912
\item gi.479140210.ref.NC\_021010.1..3168417.3168521
\item gi.238922432.ref.NC\_012781.1..c2301152.2300781
\item gi.238922432.ref.NC\_012781.1..1311347.1311667 
\item gi.238922432.ref.NC\_012781.1..c753467.753039
\item gi.238922432.ref.NC\_012781.1..c1926223.1925561
\item gi.238922432.ref.NC\_012781.1..c2743334.2742522
\item gi.479213596.ref.NC\_021044.1..1253015.1253476
\item gi.479213596.ref.NC\_021044.1..c2217719.2217459
\item gi.479208076.ref.NC\_021042.1..c657163.656546 
\item gi.545411814.ref.NZ\_KE993320.1..c60369.59749 
\item gi.211596842.ref.NZ\_DS264296.1..41162.41911 
\item gi.479140210.ref.NC\_021010.1..2564135.2565262
\item gi.238922432.ref.NC\_012781.1..c2485168.2484434
\item gi.224485442.ref.NZ\_EQ973177.1..c62806.61865
\item gi.479213596.ref.NC\_021044.1..c2624541.2623825
\end{itemize}

\newpage

\section*{Appendix D: Notation Table}

\setstretch{1.2}
\begin{longtable}{p{\dimexpr0.15\textwidth-2\tabcolsep-\arrayrulewidth\relax}|
                p{\dimexpr0.85\textwidth-2\tabcolsep-\arrayrulewidth\relax}}
  \hline
  Variable & Value / Description \\ 
  \hline
  $K$ & number of training studies \\ 
    $N_k$ & sample size in study $k$ ($k=0$ corresponds to test dataset) \\ 
    $Y_{jk}$ & outcome for individual $j$ in study $k$ \\
    $\vec{Y_k}$ & outcome vector for study $k$ \\
    $\vec{Y}$ & $(\vec{Y_1}^T, \dots, \vec{Y_K}^T)^T$; outcome vector for merged dataset\\
    $X_{ijk}$ & $i$th predictor for individual $j$ in study $k$ \\
    $\vec{X_{jk}}$ & vector of predictors for individual $j$ in study $k$ \\
    $\mat{X_k}$ & fixed effects design matrix for study $k$ \\
    $\mat{X}$  & $\begin{bmatrix}\mat{X_1}^T | \dots | \mat{X_K}^T\end{bmatrix}^T$; design matrix for merged dataset\\
    $\tilde{\mat{X_k}}$ & basis-expanded version of $\mat{X_k}$ used to fit ridge regression prediction models \\
    $\tilde{\mat{X}}$  & basis-expanded version of $\mat{X}$ used to fit ridge regression prediction models\\
    $\mat{Z_k}$  & random effects design matrix for study $k$\\
    $\mat{Z}$  & $\begin{bmatrix}\mat{Z_1}^T | \dots | \mat{Z_K}^T\end{bmatrix}^T$; random effects design matrix for merged dataset\\
     $P$ & number of original predictors (including the intercept if applicable) \\ 
    $Q$ & number of predictors with random effects \\ 
    $f_0$ & true function relating $\mat{X_{jk}}$ to $\vec{Y_{jk}}$ such that $E[\vec{Y_{jk}}] = f_0(\mat{X_{jk}})$\\
    $\vec f$ & true function relating $\mat{X_{k}}$ to $\vec{Y_{k}}$ such that $E[\vec{Y_{k}}] = f_0(\mat{X_{k}})$; $\vec f(\mat{X_{k}}) = (f_0\left(\mat{X_{1k}}\right), \dots, f_0\left(\mat{X_{N_k k}}\right))$ \\
    $\vec{\gamma_k}$ & vector of random effects \\
    $\sigma_i^2$ & variance of random effect $i$ \\
    $\mat{G}$ & $diag(\sigma_1^2, \dots, \sigma_q^2)$; covariance matrix for random effects \\ 
    $\overline{\sigma^2}$ & $tr(\mathbf{G})/P$; heterogeneity summary measure \\
    $D$ & number of unique variance values for random effects \\
    $\sigma_{(1)}^2, \dots, \sigma_{(D)}^2$ & distinct values on the diagonal of $\mat G$ \\
    $J_d$ & number of random effects with variance $\sigma_{(d)}^2$ \\
    $\vec{\epsilon_k}$ & vector of residual errors for study $k$ \\
    $\sigma_\epsilon^2$ & variance of residuals \\
    $w_k$ & ensemble weight for study $k$ \\
    $h(x)$ & basis function \\
    $\mat{I_M}^-$ & $diag(0, 1, \dots, 1)$ if an intercept is included in the ridge regression model and $\mat{I_M}$ otherwise \\
    $\vec{\hat{\beta}_k}$ & $(\mat{\tilde{X}_k}^T \mat{\tilde{X}_k} + \lambda_k \mat{I_{M}^-})^{-1}\mat{\tilde{X}_k}^T \vec{Y_k}$; ridge regression estimator from study $k$ \\
    $\vec{\hat{\beta}_{ens}}$ & $\sum_{k=1}^K w_k \vec{\hat{\beta}_{k}}$; ensemble ridge regression estimator \\
    $\vec{\hat{\beta}_{merge}}$ & $(\mat{\tilde{X}}^T \mat{\tilde{X}} + \lambda \mat{I_{M}^-})^{-1}\mat{\tilde{X}}^T \vec{Y}$; merged ridge regression estimator \\
    $\mat{\tilde {R_k}}$ & $\mat{\tilde{X}_k}^T\mat{\tilde{X}_k} + \lambda_k \mat{I_{M}^-}$ \\
    $\mat{\tilde{R}}$ & $\mat{\tilde{X}}^T\mat{\tilde{X}} + \lambda \mat{I_{M}^-}$ \\
    $\mat{\tilde{A}_k}$ & $\mat{\tilde{X}_k} \mat{\tilde{R}_k}^{-1} \mat{\tilde{X}_0}^T\mat{\tilde{X}_0} \mat{\tilde{R}_k}^{-1} \mat{\tilde{X}_k}^T$ \\
    $\mat{\tilde{C}_k}$ & $\mat{\tilde{X}_k} \mat{\tilde{R}}^{-1} \mat{\tilde{X}_0}^T \mat{\tilde{X}_0} \mat{\tilde{R}}^{-1}  \mat{\tilde{X}_k}^T$\\
    $\vec{b_{ens}}$ & $Bias(\mat{\tilde{X}_0} \vec{\hat{\beta}_{ens}} ) = \sum_{k=1}^K w_k \mat{\tilde{X}_0} \mat{\tilde{R}_k}^{-1} \mat{\tilde{X}_k}^T  \vec f(\mat{X_k}) - \vec f(\mat{X_0})$ \\
    $\vec{b_k}$ & $Bias(\mat{\tilde{X}_0} \vec{\hat{\beta}_k} )  =\mat{\tilde{X}_0} \mat{\tilde{R}_k}^{-1} \mat{\tilde{X}_k}^T  \vec f(\mat{X_k}) - \vec f(\mat{X_0})$ \\
    $\vec{b_{merge}}$ & $Bias(\mat{\tilde{X}_0} \vec{\hat{\beta}_{merge}} )  = \mat{\tilde{X}_0} \mat{\tilde{R}}^{-1} \mat{\tilde{X}}^T \vec f(\mat{X}) - \vec f(\mat{X_0})$ \\
    $\tau$ & transition point for ridge regression (Theorem 1) \\
    $\tau_{1}$ & lower bound of transition interval for ridge regression (Theorem 2) \\
    $\tau_{2}$ & upper bound of transition interval for ridge regression (Theorem 2) \\
    $v_k$ & $tr(Cov(\mat{\tilde{X}_0}  \vec{\hat{\beta}_{k}} )) = tr(\mat{Z_k} \mat{G} \mat{Z_k}^T \mat{\tilde{C}_k} + \sigma_\epsilon^2 \mat{\tilde{A}_k} )$ \\
   \hline
\caption{Notation table.}
\label{table:notation}
\end{longtable}

\end{appendices}

\end{document}